# iCORPP: Interleaved Commonsense Reasoning and Probabilistic Planning on Robots

Shiqi Zhang[a,*], Piyush Khandelwal[b] and Peter Stone[c,b]

[a]The State University of New York at Binghamton, 4400 Vestal Parkway East, Binghamton, 13902, NY, USA
[b]Sony AI, Sony Corporation of America, 25 Madison Avenue, New York, 10010, NY, USA
[c]The University of Texas at Austin, 2317 Speedway, Stop D9500, Austin, 78712, TX, USA

ABSTRACT

Robot sequential decision-making in the real world is a challenge because it requires the robots to simultaneously reason about the current world state and dynamics, while planning actions to accomplish complex tasks. On the one hand, declarative languages and reasoning algorithms support representing and reasoning with commonsense knowledge. But these algorithms are not good at planning actions toward maximizing cumulative reward over a long, unspecified horizon. On the other hand, probabilistic planning frameworks, such as Markov decision processes (MDPs) and partially observable MDPs (POMDPs), support planning to achieve long-term goals under uncertainty. But they are ill-equipped to represent or reason about knowledge that is not directly related to actions. In this article, we present an algorithm, called iCORPP, to simultaneously estimate the current world state, reason about world dynamics, and construct task-oriented controllers. In this process, robot decision-making problems are decomposed into two interdependent (smaller) subproblems that focus on reasoning to "understand the world" and planning to "achieve the goal" respectively. The developed algorithm has been implemented and evaluated both in simulation and on real robots using everyday service tasks, such as indoor navigation, and dialog management. Results show significant improvements in scalability, efficiency, and adaptiveness, compared to competitive baselines including handcrafted action policies.

## 1. Introduction

Automated reasoning and planning under uncertainty are two of the most important research areas in intelligent robotics. On the one hand, *reasoning* is concerned with using existing knowledge to efficiently and robustly draw conclusions, where the provided knowledge is typically in a declarative form. On the other hand, *planning* algorithms can be used for sequencing actions to accomplish complex tasks that require more than one action. Despite the significant achievements made in the two subareas of intelligent robotics, relatively little work has been conducted to exploit their complementary features. Focusing on applications of (semi-)autonomous robots that frequently require capabilities of both reasoning and planning, this article aims at developing a principled integration of the two computational paradigms to significantly improve robot decision-making performance in scalability, accuracy, efficiency, and adaptiveness.

In this article, we use "commonsense reasoning" to refer to rule-based methods that are able to infer not only the truthfulness of logical statements, but also their probabilities. There are many existing algorithms and corresponding systems supporting this type of logical-probabilistic reasoning (Halpern, 2017; Richardson and Domingos, 2006; Lee and Wang, 2016; Balai, Gelfond and Zhang, 2019). We use "probabilistic planning" to refer to methods that aim to compute a policy for sequential action selection based on the current (estimated) state. While also important for mobile intelligent robots, motion planning and classical automated planning methods are beyond the scope of this article. Generally, we consider domains where the current world state is partially observable, so reasoning to infer the world state is necessary. Even when we use Markov decision processes (MDPs) to build our planners, we do not assume that the whole world state is fully observable, but rather that there are aspects of the world that are not modeled in the MDPs.

This work is motivated by mobile robot platforms that have been able to navigate for unprecedented distances in recent years, while providing services such as human guidance and object delivery (Khandelwal, Zhang, Sinapov, Leonetti, Thomason, Yang, Gori, Svetlik, Khante, Lifschitz, Aggarwal, Mooney and Stone, 2017; Hawes, Burbridge, Jovan et al., 2017; Veloso, 2018; Chen, Wu, Shuai and Chen, 2017). Toward autonomy over extended periods of time, one needs the decision-making capability of simultaneously *reasoning* about the state and dynamics of the world, and *planning* to accomplish tasks. Robot decision-making has been extremely challenging, because both reasoning and planning are computationally complex problems in real-world domains: a complex robotic task frequently requires the robot to reason about a large number of objects and their properties, resulting in a high-dimensional reasoning space (the so-called "curse of dimensionality"); a robot often needs to take many actions to reach the goal of complex tasks, resulting in a long planning horizon (the so-called "curse of history") (Kurniawati, Du, Hsu and Lee, 2011).

*Corresponding author
zhangs@binghamton.edu (S. Zhang); piyush.khandelwal@sony.com (P. Khandelwal); pstone@cs.utexas.edu (P. Stone)
ORCID(s):

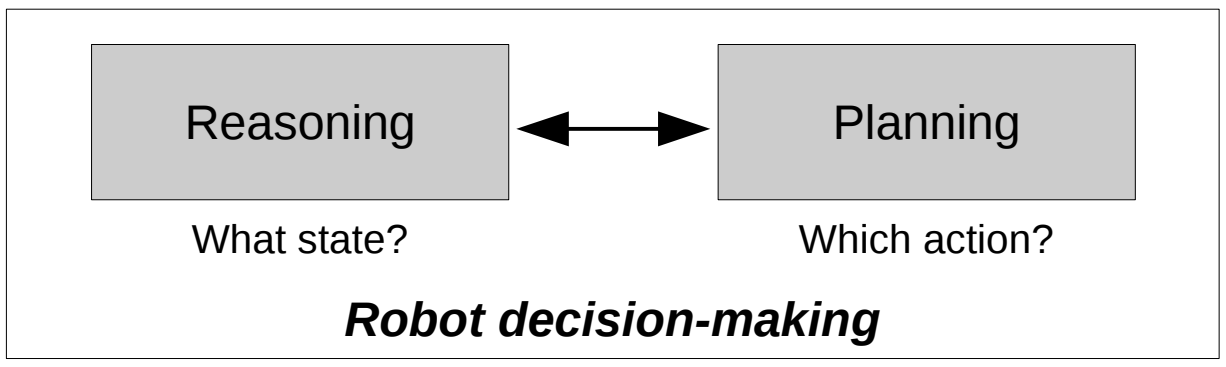

**Figure 1:** Integrated reasoning and planning (IRP) algorithms decompose a robot sequential decision-making problem into two (smaller) sub-problems that focus on reasoning about the current state of the world (including world dynamics) and selecting actions to achieve goals.

Integrated reasoning and planning (IRP) algorithms decompose a robot sequential decision-making problem into two sub-problems that focus on high-dimensional reasoning (about world state, dynamics, or both) and long-horizon planning (for goal achievement) respectively. The two interdependent sub-problems are much "smaller" than the original decision-making problems. This key idea of IRP algorithms is illustrated in Figure 1, where an IRP algorithm can be identified based on the forms of its reasoning and planning components and how they interact with each other.

In this article, we present a realization of IRP called *interleaved commonsense reasoning and probabilistic planning* (iCORPP). We build the reasoning component of iCORPP using P-log, a declarative programming paradigm that supports representing and reasoning with both logical and probabilistic knowledge (Baral, Gelfond and Rushton, 2009; Balai and Gelfond, 2017). We build the planning component of iCORPP using decision-making frameworks based on Markov Decision Processes (MDPs) or Partially Observable MDPs (POMDPs) (Puterman, 2014; Kaelbling, Littman and Cassandra, 1998), depending on the observability of the relevant aspects of the world state. Assuming a factored state space, the (single) reasoner of iCORPP faces a world model that includes all domain variables, and each planner (out of potentially many) corresponds to a partial world model that includes a minimal set of variables relevant to one task. We use the reasoner to dynamically estimate the current state of the world, and reason about the parameters of probabilistic planners (i.e., world dynamics), which are then used to construct probabilistic controllers, enabling scalable and adaptive robot decision-making.

This article builds on our previous research that appeared in two conference papers (Zhang and Stone, 2015; Zhang et al., 2017). This article unifies their terminology, problem statements, and algorithms. Compared to the prior work, this article introduces a new problem statement in Section 4.1, which covers the problems addressed in both conference papers. To address this problem, we have reformulated the iCORPP algorithm (Section 4.2) into a novel form that includes the three key steps of "logical reasoning," "probabilistic reasoning over world states," and "probabilistic reasoning about world dynamics." Despite a few new figures, the experimental results are based mainly on those from the conference papers. A new subsection in Section 6 has been introduced to discuss the applicability of iCORPP. We implement and evaluate iCORPP[1] using a mobile robot that works in an office environment on everyday service tasks of indoor navigation, and dialog management. Experimental results suggest significant improvements in both scalability and adaptiveness, in comparison to hand-coded action policies and other competitive baselines.

The remainder of this article is organized as follows. Section 2 discusses existing IRP algorithms, and how this work differs from them. Section 3 presents the two "building blocks" of this work, including P-log for logical-probabilistic knowledge representation and reasoning, and (PO)MDPs for probabilistic planning. Section 4 points to the main contribution of this article, where subsection 4.1 provides a definition of IRP problems, and subsection 4.2 describes the iCORPP algorithm. Section 5.1 summarizes the implementation strategy of iCORPP, and hypotheses used in evaluations. Sections 5.2 and 5.3 detail the implementations of iCORPP on *mobile robot navigation* and *spoken dialog system* problems respectively, where each section includes the results of evaluations using the hypotheses listed in Section 5.1. Section 6 discusses the applicability of iCORPP, and concludes this article, while listing a few open problems for future work.

## 2. Related Work

In the knowledge representation and reasoning (KRR) literature, common sense is a term that has been extensively used with different definitions – see the review article by Davis and Marcus (2015). In this article, we use the term *commonsense knowledge* to refer to the knowledge that is normally true but not always. Examples include "people prefer coffee in the mornings," and "office doors are closed over weekends." Such knowledge can be represented in a variety of forms, and we use probabilities and defaults in this article. P-log (Baral et al., 2009; Balai and Gelfond, 2017) is a declarative programming paradigm that extends Answer set programming (ASP) (Gelfond and Kahl, 2014; Lifschitz, 2008) by enabling the representation of and reasoning with probabilities. P-log and its supporting systems (Zhu, 2012; Balai, 2017) meet our need of commonsense reasoning and are used in this research. Syntax and semantics of P-log (and ASP) are summarized in Section 3.

In addition to P-log, researchers have developed many other languages and algorithms that support representing and reasoning with both logical and probabilistic knowledge, including probabilistic first-order logic (Halpern, 2017), Markov logic networks (MLN) (Richardson and Domingos, 2006), Bayesian logic (BLOG) (Milch, Marthi, Russell, Sontag, Ong and Kolobov, 2005), probabilistic Prolog (ProbLog) (De Raedt, Kimmig and Toivonen, 2007), LPMLN (Lee and Wang, 2016), and probabilistic soft logic

[1] We initially introduced a restricted version of the algorithm called CORPP (Zhang and Stone, 2015) that only reasons about world states. iCORPP in this article reasons about both states and dynamics, and CORPP is treated as an ablation that does not allow interleaving of planning and acting.

(PSL) (Kimmig, Bach, Broecheler, Huang and Getoor, 2012). These languages and algorithms were developed for different purposes, but all can be used to draw conclusions that are associated with probabilities. Most of these computational paradigms are concerned with a static world, meaning that the programs do not look into world changes over time, while some can be used for modeling planning domains (Eiter and Lukasiewicz, 2002; Zhu, 2012; Lee and Wang, 2018) as optimization problems to find actions leading to the goal state *with the highest probability*. Despite the KRR strengths of these languages and algorithms, none of these (including P-log) support planning under uncertainty toward maximizing cumulative reward over a long, unspecified horizon, which is frequently required while a robot is working on complex tasks. pBC+ is a KRR paradigm that can be used for declaratively encoding MDPs and POMDPs (Wang, Zhang and Lee, 2019), and further supports the automatic construction of (PO)MDP programs that can be processed by systems for planning under uncertainty. pBC+ can be used to realize the idea described in this article.

Within the context of AI, there are mainly two classes of planning algorithms, fully observable deterministic planning (frequently referred to as classical planning) and planning under uncertainty.[2] Classical planning algorithms, e.g., Fast Forward (FF) (Hoffmann and Nebel, 2001) and Fast Downward (FD) (Helmert, 2006), focus on computing a sequence of actions, implicitly assuming perfect action executions in a deterministic domain. Classical planning domains and problems are usually formalized using *action languages* (Fikes and Nilsson, 1971; McDermott, Ghallab, Howe, Knoblock, Ram, Veloso, Weld and Wilkins, 1998; Lee, Lifschitz and Yang, 2013), where early action languages are surveyed in (Gelfond and Lifschitz, 1998). Algorithms for planning under uncertainty, e.g., Value Iteration (Sutton and Barto, 2018) and UCT (Kocsis and Szepesvári, 2006), aim at computing an action policy that suggests an action from any state under the uncertainty from the non-deterministic outcomes of robot actions. Examples of non-deterministic action outcomes include opponent moves in chess and results of grasping an object using an unreliable gripper. This article focuses on the class of planning under uncertainty in stochastic domains, although the developed algorithms have potential applications to both classes.

Sequential decision-making frameworks, such as MDPs (Puterman, 2014) and partially observable MDPs (POMDPs) (Kaelbling et al., 1998), can be used for planning under uncertainty toward maximizing long-term reward. These frameworks and their descriptive languages, such as PPDDL (Younes and Littman, 2004) and RDDL (Sanner, 2010), well support the representation of and reasoning about action knowledge. However, they are not designed for, and are hence less effective in, tasks that require reasoning about stationary worlds, e.g., to query whether a state is valid or estimate the current state of the world *before* taking any action. Logical-probabilistic KRR languages, such as P-log, are suitable for such reasoning tasks. In short, both the commonsense reasoning and probabilistic planning paradigms have strengths and weaknesses.

As a result, integrated reasoning and planning (IRP) algorithms have been developed in recent years (Zhang and Sridharan, 2022). For instance, logical reasoning has been incorporated into planning under uncertainty to compute informative prior distributions (Zhang, Sridharan and Wyatt, 2015a), where domain-dependent heuristics are required to generate such priors, limiting its applicability to complex problems. The OpenDial system integrates probabilistic reasoning and POMDP-based probabilistic planning (Lison and Kennington, 2016), but their approach was specifically developed for the application of dialog management, limiting the applicability to other domains. A two-level, refinement-based architecture has been developed for robot reasoning and planning (Sridharan, Gelfond, Zhang and Wyatt, 2019). The high-level reasoner is used for computing a deterministic sequence of actions to guide a low-level probabilistic controller. The reasoner also supports complicated reasoning tasks, such as explaining past behaviors, that are impossible for probabilistic planners. In the work of Hanheide, Göbelbecker, Horn, Pronobis, Sjöö, Aydemir, Jensfelt, Gretton, Dearden, Janicek, Zender, Kruijff, Hawes and Wyatt (2017), commonsense reasoning was used for diagnostic tasks and generating explanations, and a hybrid planner allows switching between deterministic and probabilistic planners. Human-provided information is provided to probabilistic controllers in domains with probabilistic relational constraints (Chitnis, Kaelbling and Lozano-Perez, 2018). Unlike the above algorithms, iCORPP uses a logical-probabilistic paradigm for KRR, and allows dynamically reasoning about and constructing complete, probabilistic controllers.

Algorithms have been developed for integrating reinforcement learning (RL) (Sutton and Barto, 2018) and commonsense reasoning. Leonetti, Iocchi and Stone (2016) used action knowledge to help a robot select reasonable actions in exploration. In that work, the robot used a RL algorithm to learn an action policy in unknown environments. Sridharan, Meadows and Gomez (2017) used relational RL to learn action affordances that are needed by a reasoner for both reasoning and planning tasks. Focusing on non-stationary domains, reasoning methods have been used to help find possible trajectories for reinforcement learners (Ferreira, Bianchi, Santos and de Mantaras, 2017). Declarative action knowledge has been integrated with hierarchical RL for improving the learning rate of a reinforcement learner (Yang, Lyu, Liu and Gustafson, 2018; Jiang, Yang, Zhang and Stone, 2019). Reward machines from domain experts for encoding temporal logics have been used to guide RL agents to learn faster via exploiting the internal structures of reward functions (Icarte, Klassen, Valenzano and McIlraith,

[2]The broadly defined planning problem also includes motion planning (LaValle, 2006) that focuses on computing trajectories in continuous space. Motion planning, and integrated task and motion planning (TAMP) (Garrett, Chitnis, Holladay, Kim, Silver, Kaelbling and Lozano-Pérez, 2021) are beyond the scope of this article.

2018, 2020). In our recent work, model-based RL was used for learning world dynamics, where a robot uses the learned knowledge to construct probabilistic controllers on the fly, while accounting for new circumstances (Lu, Zhang, Stone and Chen, 2020). In these works, machine learning algorithms (RL in particular) were used for improving the reasoning or planning components of IRP methods. The above algorithms (integrating reasoning and RL) can potentially be applied to IRP algorithms to enable the planning components to evolve over time and experience, though learning is not a focus in this article. iCORPP has the potential to enable promising lines of research that involve reasoning, planning, and learning, as discussed in our future work (Section 6).

The strategy of decomposing sequential decision-making tasks into reasoning and planning has been observed in the study of human behaviors (Bazerman and Moore, 2008; Triantaphyllou, 2000). For instance, existing research on human decision-making has provided empirical evidence that people make decisions by first understanding the domain (including analyzing a discrete set of alternatives) and then finding the optimal solution (by evaluating the impact of the alternatives on certain criteria) to maximize the overall utility (Triantaphyllou, 2000). From the perspective of decomposing decision-making tasks into reasoning and planning subtasks, iCORPP functions like the process of human decision-making, as evidenced by human behavior research (Bazerman and Moore, 2008), and can potentially be used for the imitation of human decision-making processes.

Finally, we qualitatively compare this article with three related conference papers, upon two of which it is based. Compared with CORPP (Zhang and Stone, 2015), this article supports reasoning about beliefs, dynamics, and rewards of a planning system, whereas CORPP only supports reasoning about beliefs. Compared with iCORPP (Zhang et al., 2017), this article includes a problem statement that describes the input, output, and assumptions of our approach. Additionally, this article unifies the terminology of the CORPP and iCORPP papers (see Section 4.1), and reformulates the algorithm accordingly (see Section 4.2). These changes required a significant amount of detail-oriented thinking and re-writing. Finally, PBCPLUS2POMDP (Wang et al., 2019) is a system paper, which developed a realization of the approach described in the original iCORPP paper (Zhang et al., 2017). The PBCPLUS2POMDP system was realized using a knowledge representation and reasoning paradigm called PBCPLUS, whereas this article focuses on formally presenting the problem and an algorithm for addressing the problem. This article is not based on the PBCPLUS2POMDP system.

## 3. Background

In this section, we review the substrate techniques used in this article for knowledge representation and reasoning (KRR) and planning under uncertainty respectively. Specifically, we use P-log (Baral et al., 2009) for KRR, and use Markov decision processes (MDPs) (Puterman, 2014) and partially observable MDPs (POMDPs) (Kaelbling et al., 1998) for planning under uncertainty.

### 3.1. Answer Set Programming and P-log

Answer set programming (ASP) (Baral, 2003; Gelfond and Kahl, 2014) is a non-monotonic logic programming paradigm with stable model semantics (Gelfond and Lifschitz, 1988). ASP has been applied to a variety of problem domains (Erdem, Gelfond and Leone, 2016), including robotics (Erdem and Patoglu, 2018).

An ASP program can be described as a five-tuple $\langle \Theta, \mathcal{O}, \mathcal{F}, \mathcal{P}, \mathcal{V} \rangle$ of sets. These sets contain names of the *sorts*, *objects*, *functions*, *predicates*, and *variables* used in the program, respectively. Variables and object constants are *terms*. An *atom* is an expression of the form `p(t)=true` or `a(t)=y`, where `p` is a predicate, `a` is a function, `y` is a constant from the range of `a` or a variable, and `t` is a vector of terms. For example, `alice` is an object of sort `person`. We can define a predicate `prof` and use `prof(P)` to identify whether person `P` is a professor, where `P` is a variable.

A *literal* is an atom or its negation, where an atom's negation is of the form `p(t)=false` or `a(t)`$\neq$ `y`. In this article, we call `p(t)` and `a(t)` *attributes*, if there is no variable in `t`. For instance, `prof(alice)=true` is a literal and we can say the value of attribute `prof(alice)` is `true`. For simplicity's sake, we replace `p(t)=true` with `p(t)` and `p(t)=false` with `-p(t)` in the rest of this article.

An ASP program consists of a set of rules of the form:

```
p, ..., q :- r, ..., s, not t, ..., not u.
```

where `{p,...,u}` are literals, "`:-`" is a Prolog-style implication sign, and symbol `not` is a logical connective called *default negation*.

A rule is separated by the symbol "`:-`", where the left side is called the *head* and the right side is called the *body*. A rule is read as "head is true if body is true". A rule with an empty body is referred to as a *fact*. If `l` is a literal, expressions `l` and `not l` are called *extended literals*. Default negation supports reasoning about unknowns, and `not l` is read as "it is unknown that `l` is true", which does not imply that `l` is believed to be false. For instance, `not prof(alice)` means it is not believed that `alice` is a professor or there is no evidence supporting `alice` being a professor.

Using default negation, ASP can represent (prioritized) default knowledge with different levels of exceptions. Default knowledge allows us to draw tentative conclusions by reasoning with incomplete information and commonsense knowledge. The rule below shows a simplified form of defaults that only allows *strong exceptions* that refute the default's conclusion: for object X of property c, it is believed that X has property p, if there is no evidence to the contrary.

$$\mathtt{p(X)} \leftarrow \mathtt{c(X)},\ \mathtt{not}\ \neg\mathtt{p(X)}.$$

Traditionally, ASP does not explicitly quantify degrees of uncertainty: a literal is either true, false or unknown. P-log (Baral et al., 2009) is an extension to ASP that allows *random selections*. A random selection states that, if `B` holds, the value of `a(t)` is selected randomly from the set `{X:q(X)} ∩ range(a)`, unless this value is fixed elsewhere:

```
random(a(t): {X:q(X)}) :- B.
```

where `B` is a collection of extended literals; `q` is a predicate.

Finally, the following *probability atom* (or *pr-atom*) states that, if `B` holds, the probability of `a(t)=y` is `v` $\in [0, 1]$.

```
pr(a(t)=y|B)=v.
```

Reasoning with an ASP program generates a set of *possible worlds*: $\{W_0, W_1, \cdots\}$, where each is in the form of an answer set that includes a set of literals. Probabilistic models consist of a finite set whose elements are referred to as possible worlds (Gelfond and Kahl, 2014). The random selections and pr-atoms enable P-log to calculate a probability for each possible world. Therefore, ASP and P-log together enable one to draw inferences regarding possible (and impossible) world states using the strong capabilities of representing and reasoning with (logical and probabilistic) commonsense knowledge. P-log systems, such as those developed by Zhu (2012) and Balai et al. (2019), use causal Bayesian networks (Pearl, 2014) to compute a probability for each possible world. Neither ASP nor P-log supports planning under uncertainty toward maximizing long-term rewards with long, unspecified horizons.

### 3.2. MDPs and Partially Observable MDPs

The Markov property states that the next state depends on only the current state and action, and is independent of all previous states and actions (the first-order case). Following the Markov assumption, a Markov decision process (MDP) can be described as a four-tuple $\langle \mathcal{S}, \mathcal{A}, T, R \rangle$. $\mathcal{S}$ defines all possible states of the world. In this article, we assume a factored state space, where a state can be specified using a set of attributes and their values. $\mathcal{A}$ is a set of actions, where an action leads state transitions by changing the value(s) of domain attribute(s); $T : \mathcal{S} \times \mathcal{A} \times \mathcal{S} \rightarrow [0, 1]$ represents the probabilistic state transition; and $R : \mathcal{S} \times \mathcal{A} \rightarrow \mathbb{R}$ specifies the rewards. Solving an MDP produces a *policy* $\pi : s \mapsto a$ that maps the current state $s$ to action $a$ in such a way that maximizes long-term rewards.

A POMDP generalizes a MDP by assuming the partial observability of the current state. As a result, a POMDP can be described as a six-tuple $\langle \mathcal{S}, \mathcal{A}, T, Z, O, R \rangle$, where $Z$ is a set of observations; $O : \mathcal{S} \times \mathcal{A} \times Z \rightarrow [0, 1]$ is the observation function; and the definitions of $\mathcal{S}$, $\mathcal{A}$, $T$, and $R$ are inherited from MDP. Unlike MDPs, the current state can only be estimated through observations in POMDPs. A POMDP hence maintains a *belief state* (or simply *belief*), $b$, in the form of a probability distribution over all possible states.

The belief update of a POMDP proceeds as follows:

$$b'(s') = \frac{O(s', a, o) \sum_{s \in \mathcal{S}} T(s, a, s') b(s)}{pr(o|a, b)} \quad (1)$$

where $s$, $a$ and $o$ represent a state, an action and an observation respectively; and $pr(o|a, b)$ is a normalizer. Solving a POMDP produces a *policy* $\pi : b \mapsto a$ that maps beliefs to actions in such a way that maximizes long-term rewards.

MDPs and POMDPs enable principled decision making under uncertainty, but are ill-equipped to scale to large numbers of domain variables or reason with commonsense knowledge that is not directly relevant to actions. Intuitively, we use ASP and P-log to represent the commonsense knowledge that includes all domain attributes, and use MDPs and POMDPs to model a subset of attributes that are needed for computing the action policy for a specific task. Therefore, given a task, there can be many of the attributes that contribute to calculating the POMDP priors, parameters, or both. Section 4.2 describes the technical details of using a commonsense reasoner to reason about world state and dynamics, and dynamically construct MDP and POMDP probabilistic controllers.

Existing work has investigated modeling exogenous events, e.g., sunlight reduces success rate of a robot navigating through an area (due to the limitations of rangefinder sensors), *within* decision-theoretic models (Boutilier, Dean and Hanks, 1999; Hoey, Poupart, Boutilier and Mihailidis, 2014). However, it is often difficult to predict how an exogenous change will affect the system state, and what the distribution for the occurrence of these exogenous events will be. Doing so also presents a trade-off between model correctness and computational tractability (as more domain variables are modeled). Although it is possible to implement domain-specific planners to efficiently handle the exogenous events, we argue that, from a practical perspective, using commonsense reasoning to shield exogenous domain attributes from MDPs and POMDPs is relatively a much more easy-to-use approach than directly manipulating probabilistic controllers' graphical representations.

## 4. Our Approach

In this section, we first present a formal statement about the domains and problems that we are interested in, and then describe our algorithm for addressing the problem.

### 4.1. Problem Statement

In this subsection, we first categorize domain variables, then motivate the problem statement, and finally present the technical problem definition.

***Categorization of Domain Variables:*** An *integrated reasoning and planning* (IRP) problem has a factored state space that is specified by a finite set of variables, $\mathcal{V}$.

- **Endogenous** variables, $\mathcal{V}^{en} \subseteq \mathcal{V}$, are the variables whose values the robot *actively* changes, estimates probabilistically, or both. Note that for this purpose, using logical reasoning to determine the value of a variable is not considered probabilistic estimation, and thus does not render a variable endogenous.
- **Exogenous** variables, $\mathcal{V}^{ex} \subseteq \mathcal{V}$, are the variables whose values the robot observes and adapts to (Boutilier et al., 1999). Exogenous variables are the ones that can be excluded from the (PO)MDP the agent needs to solve when determining its action policy.

An endogenous (exogenous) **event** corresponds to an update of an endogenous (exogenous) variable. $\mathcal{V}^{en} \cap \mathcal{V}^{ex}$ is empty, and $\mathcal{V} = \mathcal{V}^{en} \cup \mathcal{V}^{ex}$ specifies a factored state space $S_{\mathcal{V}}$. It will not harm correctness if more variables are designated as endogenous. But in practice, one should designate as many variables as possible as being exogenous, so as to limit the computational complexity of finding a good behavior policy. $A$ is a finite set of actions that a robot can perform.

*Example 1:* Consider a 2D navigation domain, where a robot knows its current position, and needs to navigate to a goal using actions that only probabilistically lead the robot to its desired positions. The $x$-axis and $y$-axis positions are endogenous variables, because the robot wants to actively change their values. One can formulate this problem using a standard MDP. Now we further consider that positions near windows are under sunlight (which probabilistically blinds the range sensors) when there are no clouds – the default case. In this extended 2D navigation problem, the "cloud" variable is exogenous, because the robot cannot change its value and does not actively estimate it either – its value can be directly observed. Additionally, for each position, there are two exogenous variables indicating its "near window" and "sunlit" statuses.

*Example 2:* Consider the partially observable tiger problem introduced by Kaelbling et al. (1998), where a tiger lurks behind one of two doors. A robot can use two "listen" actions for estimating behind which door the tiger is, and take an "open door" action to produce a big reward (if no tiger behind the door) or a big penalty (otherwise). The tiger's real position is the only endogenous variable, because the robot cannot directly observe it, but has to make observations to actively estimate its value (though it cannot be changed). Note that a variable is endogenous if a robot actively changes its value *or* probabilistically estimate it. The perceived tiger position is an exogenous variable, because it can be observed (even though its value is available only after a "listen" action) but not actively changed. In other words, the perceived position serves as fully observable "evidence", which is exogenous and can be used for updating the robot's endogenous belief about the tiger's position. In our extended tiger problem, let us further consider that the tiger's position is more perceivable when it is awake. It is known that the tiger sleeps at night when its monkey neighbor does not scream. The monkey's behavior is out of the robot's control, but fully observable. The "awake," "time," and "screaming" statuses are exogenous variables. The robot adapts its behaviors to their values, which are either observable or can be logically inferred.

> *Intuitively, one can reason about the exogenous variables to infer the values of endogenous variables, whose values are needed for planning. This categorization of variables is the key idea of this article.*

***State Knowledge and Action Knowledge:*** State knowledge $\mathcal{K}^S$ is in the form of a set of logical and probabilistic rules about $\mathcal{V}$, and implicitly specifies a *causal* Bayesian network (CBN), where each node corresponds to a single state $s \in S_{\mathcal{V}}$ that represents one of the combinatorial possible settings of the variables in $\mathcal{V}$. A state is equivalent to a *possible world* using the terminology of logic programming. $\mathcal{K}^S$ includes commonsense knowledge that is *normally* true but not always. In each possible world, it is guaranteed that each $v^{ex} \in \mathcal{V}^{ex}$ has a value that is either directly encoded in $\mathcal{K}^S$ as a default value, or can be inferred using other variables' default values. Action knowledge $\mathcal{K}^A$ is in the form of a set of logical and probabilistic rules about $\mathcal{V}$ and $A$, and, together with $\mathcal{K}^S$, implicitly specifies one *dynamic* Bayesian network (DBN) for each $a \in A$. A DBN is a Bayesian network which relates variables to each other over adjacent time steps. The DBNs share the same set of nodes (variables) inherited from the above-mentioned CBN.

***Description of an IRP Domain:*** An IRP domain includes a finite set of exogenous variables $\mathcal{V}^{ex}$, a finite set of endogenous variables $\mathcal{V}^{en}$, a finite set of actions $A$, state knowledge $\mathcal{K}^S$, action knowledge $\mathcal{K}^A$, and reward function $R : S_{\mathcal{V}} \times A \rightarrow \mathbb{R}$. To summarize, a domain description is a tuple:

$$\mathcal{D} = \langle \mathcal{V}^{ex}, \mathcal{V}^{en}, A, \mathcal{K}^S, \mathcal{K}^A, R \rangle$$

Latent variables are variables that are not directly observed but are rather inferred from other variables that are observed. If there exists at least one endogenous variable $v \in \mathcal{V}^{en}$ that is a latent variable, then we say the current world state of the domain is *partially observable*. Otherwise, the current world state is *fully observable*. We consider that the properties of exogenous variables do not affect the observability of IRP domains, because reasoning with defaults (Reiter, 1980) enables a robot to use assumed values to reason about exogenous variables. From the perspective of sequential decision-making, the values of exogenous variables are fully observable and defeasible by subsequent observations. For partially observable domains, the agent maintains a belief distribution $b$ over $S_{\mathcal{V}}$.

***Input of an IRP Algorithm:*** The input of an IRP algorithm includes a set of logical rules, a set of probabilistic rules, and a reward function $R$. The rules are used for specifying $\mathcal{V}^{ex}$, $\mathcal{V}^{en}$, $A$, $\mathcal{K}^S$, and $\mathcal{K}^A$.

Logical rules are used for specifying exogenous variables $\mathcal{V}^{ex}$ and their logical relations within an IRP problem. Each logical rule is of the form:

```
p :- r1, r2, ..., not u1, not u2, ....
```

where `{p,r1,r2,...,u1,u2,...}` are literals, and each literal is an atom or its negation. An *atom* is an expression of the form `f(t)=y`, where `f` is a function, `y` is a constant or a schematic variable, and `t` is a vector of terms. An **exogenous variable** $v^{ex}_{f(t)} \in \mathcal{V}^{ex}$ is defined as `f(t)`. The domain of each $v^{ex}_{f(t)}$ may be provided; otherwise, a P-log system (Zhu, 2012; Balai et al., 2019) assigns a default domain of `{true, false}`.

Probabilistic rules, in the form of pr-atoms in P-log, are used for specifying **endogenous variables** $\mathcal{V}^{en}$, and their probabilistic relations that are conditioned on the values of $\mathcal{V}^{ex}$. The following pr-atom about $\mathcal{V}^{en}_{g(t)}$ states that, if `B` holds, the probability of `g(t)=y` is `v`:

```
pr(g(t)=y|B)=v.
```

where `B` is a collection of extended literals about $\mathcal{V}^{en}$ and $\mathcal{V}^{ex}$. An extended literal `l` is a literal or its default negation. A conditional probability table may be provided for each endogenous variable; otherwise, P-log systems assume uniform distributions. The above-mentioned logical and probabilistic rules together specify $\mathcal{V}^{ex}$ and $\mathcal{V}^{en}$ of domain $\mathcal{D}$, as well as its **state knowledge** $\mathcal{K}^S$.

Action set $A$ and action knowledge $\mathcal{K}^A$ are also defined as pr-atoms. Action $a \in A$ is defined as a P-log random variable, `action`. Here we use different font styles to distinguish "actions" in P-log rules and IRP problem definitions. $\mathcal{K}^A$ is defined using pr-atoms in the following form:

```
pr(f'(t)=y'|f(t)=y, action=a, r1, r2, ...,
                     not u1, not u2,...)=v.
```

where `f'(t)` under a different function name is a duplicate of `f(t)`. The above rule states that action `a` changes the value of endogenous variable $\mathcal{V}^{en}_{f(t)}$ from `y` to `y'` with probability `v` in the presence of extended literals `r1, r2, ..., not u1, not u2,...`. Such rules are used for representing action knowledge $\mathcal{K}^A$.

Reward function $R$ is defined by enumerating the rewards given the values of variables and an action.

$$R(v^{en}_0 = V^{en}_0, v^{en}_1 = V^{en}_1, \cdots, \\ v^{ex}_0 = V^{ex}_0, v^{ex}_1 = V^{ex}_1, \cdots, a) = r$$

Next we use Example 1 (Navigation under Sunlight) to illustrate the input of IRP algorithms. The endogenous variables in this domain include only the two for representing the robot's x and y coordinates, because the robot only needs to change the values of those two variables for navigation purposes. To reason about the state space, there are exogenous variables about current time, and whether each grid cell is near a window or under sunlight. All those endogenous and exogenous variables are provided as part of the input. There are logical and probabilistic rules provided to the agent. For instance, one rule is that a grid cell, specified by its x and y coordinates (endogenous variables), is under sunlight, if it is next to a window (an exogenous variable) and the current time is morning (another exogenous variable). Such rules are provided as part of the input. The actions include the NESW directions of North, East, South and West. There are probabilistic rules provided as prior knowledge. One example is that a robot's navigation action fails with probability 0.9 when it navigates under sunlight; otherwise, a navigation action leads to its intended location. Such rules about world dynamics are provided as part of the input. Finally, the robot's navigation goal is specified by giving it a big reward (say 100) if the robot lands in one of the goal cells.

***Output of an IRP Algorithm:*** The output of an IRP algorithm is an action policy $\pi$. The objective is to compute $\pi$ for the robot to choose actions at each time step toward maximizing its expected future discounted reward,

$$\pi^* = \operatorname*{argmax}_{\pi} E\left[\sum_{t=0}^{\infty} \gamma^t r_t\right]$$

where $\gamma$ is a discount factor that determines how much immediate rewards are favored over more distant rewards, and $r_t$ is the reward received at time $t$.

***Assumptions:*** We make the following set of assumptions about IRP domains:

- The program that consists of the logical rules about exogenous variables is satisfiable, i.e., there is no inconsistency among the provided logical statements.
- Values of exogenous variables are either fully observable or unobservable. When fully observable, they are formulated as logical facts; otherwise, their values can be inferred using the other variables.
- An exogenous variable's value can become known during the execution of a policy, and an IRP agent is immediately aware of such changes.
- An endogenous variable is either partially observable or fully observable. An endogenous variable's observability does not change over time, and is known.

***Remarks:*** One might wonder why there is no observation set defined in $\mathcal{D}$ in case of partially observable domains. The reason is that the DBNs, one for each $a \in A$, already allow updating beliefs based on the current belief, values of exogenous variables (or evidence variables), and the performed action. In other words, the observation function can be inferred from the DBNs specified by $\mathcal{D}$.

Figure 2 illustrates the high-level structure of partially observable IRP problems. In standard POMDPs (Kaelbling et al., 1998), observations are active, because what an agent observes depends on the performed action. In partially observable IRP domains, each observation (modeled in $\mathcal{V}^{ex}$) includes two components: active and passive. The component of observations for estimating $\mathcal{V}^{en}$ is active. For instance, the tiger's perceived position belongs to the active component of observations, because it is used for estimating the tiger's real position. $\mathcal{V}^{ex}$ also models the passive component of observations (e.g., the perceived "screaming" status) that are not affected by an agent's actions. The passively observed component of $\mathcal{V}^{ex}$ can potentially affect the agent's action policy, whereas the active components cannot. This main difference from the standard POMDP diagram is marked using a red dashed line in Figure 2. It should be noted that the introduction of IRP observations' active and passive components is only for the comparison with the standard POMDP's diagram, and is not required by IRP algorithms. The realization of state estimator (SE) is the same as that of standard POMDPs, as summarized in Section 3.

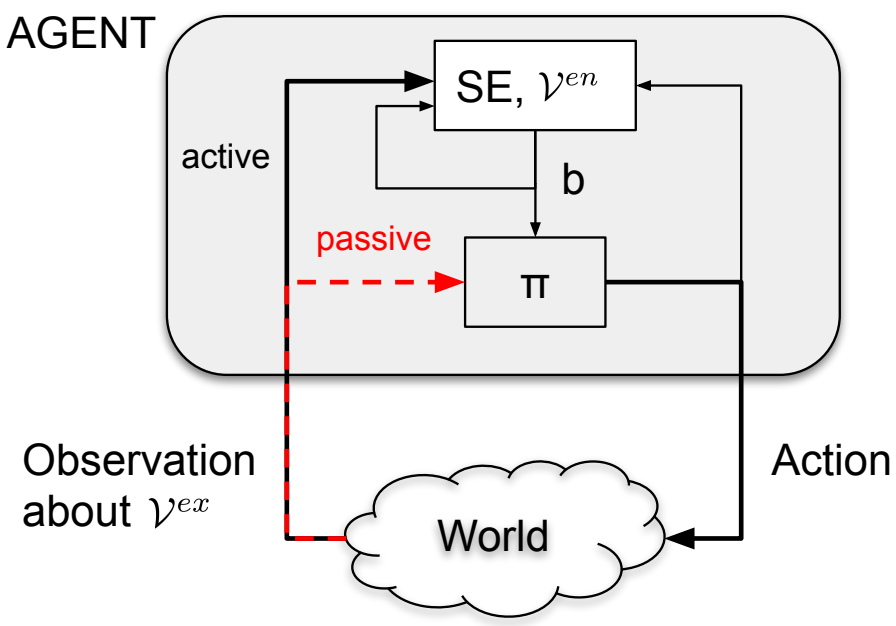

**Figure 2:** Diagram of integrated reasoning and planning (IRP) problems that includes two main components: *State Estimator* (SE) for updating belief $b$, and policy $\pi$ for action selection based on $b$. Policy $\pi$ is computed using domain description $\mathcal{D}$ and the observation over exogenous variables.

Generally, in partially observable IRP domains, exogenous variables affect beliefs over endogenous variables. The policy maps beliefs over endogenous variables to actions. So while the exogenous variables indirectly affect the policy via the endogenous variables, only the endogenous variables are actually inputs to the policy. Next, we describe our approach for addressing IRP problems.

## 4.2. iCORPP, Our Approach

The key idea of this article is to reason with declarative logical-probabilistic knowledge about the world state and dynamics for planning under uncertainty. Figure 3 illustrates the interleaved commonsense reasoning and probabilistic planning (iCORPP) process. In this section, we discuss the following topics:

(I) Using logical reasoning to specify a task-oriented partial state space;
(II) Probabilistic reasoning for computing a belief distribution over states (in case of domains under partial observability); and
(III) Probabilistic reasoning about transition function that may change over time, i.e., world dynamics.

The three steps together enable an agent to dynamically construct probabilistic graphical models on the fly via reasoning with logical-probabilistic commonsense knowledge. After that, reward functions, as well as observation functions in case of partially observable domains, can be constructed accordingly to form complete (PO)MDPs. Algorithms for planning under uncertainty take the (PO)MDP models as input, and generate action policies.

***iCORPP, the Algorithm:*** iCORPP is described in Algorithm 1. We use off-the-shelf reasoning and planning systems, including $Sol^{lr}$ for logical reasoning, $Sol^{pr}$ for probabilistic reasoning, and $Sol^{pl}$ for planning under uncertainty. $Sol^{lr}$ takes as inputs Prolog-style logical statements about variables $\mathcal{V}$, and outputs a set of possible worlds $\mathcal{W}$, where $w \in \mathcal{W}$ gives each variable a value. The input statements can potentially include commonsense statements that are not always correct. $Sol^{pr}$ takes logical statements (same as those provided to $Sol^{lr}$) and probabilistic statements as the input, and computes a probability for each possible world. It should be noted that, in our implementation of iCORPP, the reasoning system we use supports the functionalities of both $Sol^{lr}$ and $Sol^{pr}$, while we separate the two systems to discuss the general case. Without actions in the input of $Sol^{pr}$, each possible world corresponds to a state in (PO)MDP terminology. When action descriptions are provided to $Sol^{pr}$, one can build the correspondence between possible worlds and state-action pairs.

Here we look into the steps of iCORPP as presented in Algorithm 1. Entering the main loop of Steps 1-19, in Steps 2-3, the robot collects facts $\mathcal{F}$ over $\hat{\mathcal{V}}^{ex} \subseteq \mathcal{V}^{ex}$, fully observable exogenous variables. Step 4 is for logical reasoning to compute a set of possible worlds $\mathcal{W}^{cplt}$, where $w^{cplt} \in \mathcal{W}^{cplt}$ corresponds to a complete assignment to all variables (endogenous and exogenous). Dimensionality reduction in Step 5 filters out the exogenous variables to compute a smaller state space $\mathcal{W}$ focusing on $\mathcal{V}^{en}$.[3] Step 6 bridges the gap between logic programming, and planning under uncertainty by building the correspondence between state $s \in S$ and $w \in \mathcal{W}$. $\mathcal{W}^{cplt}$ and $\mathcal{W}$ (or $S$) correspond to the complete state space and partial state space respectively, as shown in Figure 3.

Step 7 evaluates the state observability, and determines if it is necessary to maintain a belief distribution for state estimation. If so, Step 8 calls a probabilistic reasoning system to compute the probability of each possible world, which corresponds to "Belief state (task-oriented)" in Figure 3. After that, we assume the robot's perception model is provided (though it can be learned from data), and hence the observation function $\mathcal{O}$ can be constructed in Step 9.

Step 11 calls a probabilistic reasoning system to compute the transition function. This is a rather complex process; we discuss the implementation of Step 11 after describing iCORPP in general. In Step 12, iCORPP constructs (PO)MDP models. Steps 13 is for policy generation. For instance, $Sol^{pl}$ can compute a policy $\pi : s \rightarrow a$ using algorithms such as SARSOP (for POMDPs) (Kurniawati, Hsu and Lee, 2008) and value iteration or Monte Carlo tree search (for MDPs) (Kocsis and Szepesvári, 2006).

Steps 14-18 are for interaction with the real world using the generated policy. To account for exogenous events at execution time, Step 17 updates facts $\mathcal{F}$ in each iteration,

[3] The "dimensionality reduction" referred to in Step 5 is also called "abstraction" in the AI literature. In this article, the dimensionality reduction process relies on expert knowledge that specifies the relevance between variables and tasks. We leave the automation of this process to future work.

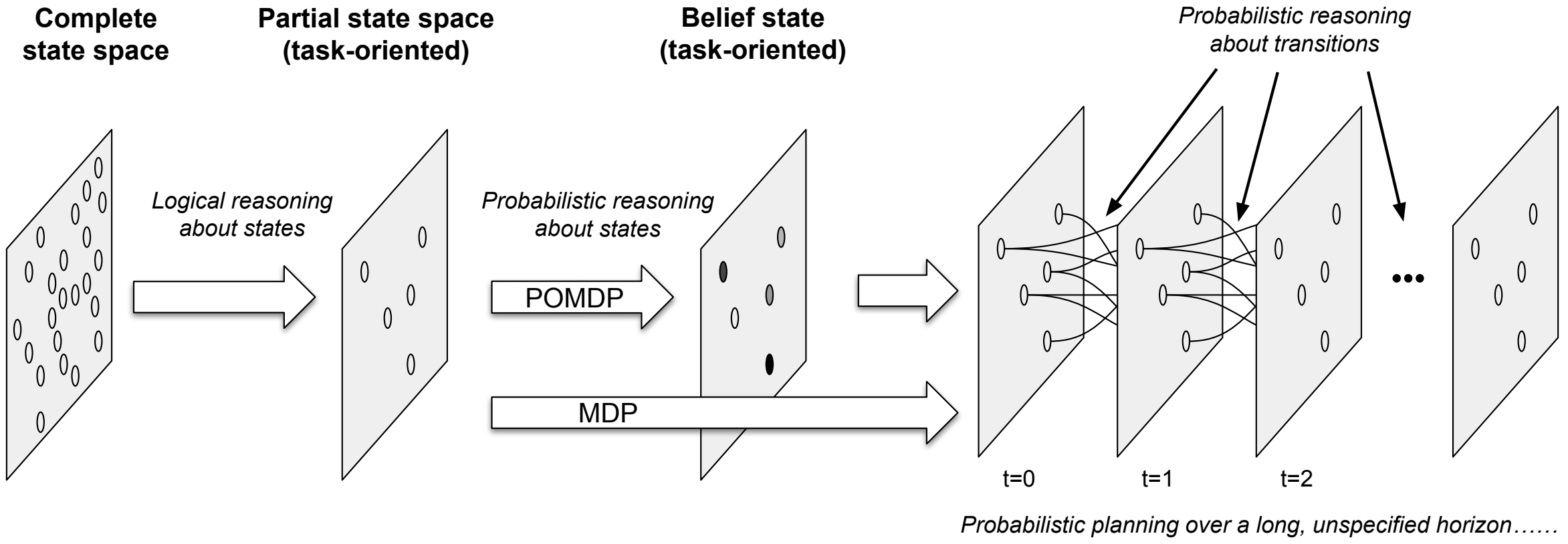

**Figure 3:** Overview of iCORPP. The original (complete) state space includes potentially many states. The logical-reasoning step produces a partial state space through a dimensionality-reduction process, focusing on the endogenous variables required by the current task. In case of the current state being partially observable, iCORPP computes a prior belief distribution over the states. Finally, iCORPP reasons about world dynamics for task-oriented planning under uncertainty.

---

**Algorithm 1** iCORPP

---

**Input:** $\mathcal{V}^{ex}$; $\mathcal{V}^{en}$; $A$; $\mathcal{K}^S$; $\mathcal{K}^A$; $R$.
**Require:** solver for logical reasoning $Sol^{lr}$; solver for probabilistic reasoning $Sol^{pr}$; solver for planning $Sol^{pl}$

1: **repeat**
2:   Iterate over $\mathcal{V}^{ex}$, identify the fully observable subset $\hat{\mathcal{V}}^{ex} \subseteq \mathcal{V}^{ex}$, and collect values of $\hat{\mathcal{V}}^{ex}$
3:   Convert assignments of $\hat{\mathcal{V}}^{ex}$ into a logical form, $\mathcal{F}$, referred to as facts
4:   $\mathcal{W}^{cplt} \leftarrow Sol^{lr}(\mathcal{F}, \mathcal{K}^S)$, where $w^{cplt} \in \mathcal{W}^{cplt}$ is a complete assignment to variables $\mathcal{V}^{en} \cup \mathcal{V}^{ex}$ // Logical reasoning
5:   Generate $\mathcal{W}$ using $\mathcal{W}^{cplt}$ (dimensionality reduction), where $w \in \mathcal{W}$ is an assignment to $\mathcal{V}^{en}$
6:   Construct state space $S$, where $s \in S$ corresponds to $w \in \mathcal{W}$
7:   **if** $\exists V^{en} \in \mathcal{V}^{en}$ that is latent **then**
8:     Call $Sol^{pr}(\mathcal{W}, \mathcal{K}^S)$ to compute $Pr(w)$ for each $w \in \mathcal{W}$; $b(s) \leftarrow Pr(w)$ // Probabilistic reasoning about states
9:     Generate observation function $\mathcal{O}$ accordingly: $\mathcal{O}(o, s', a) = Pr(o|s', a)$
10:   **end if**
11:   Call $Sol^{pr}((\mathcal{W}, A), \mathcal{K}^A)$ to compute $T(s, a, s')$ // Probabilistic reasoning about transitions
12:   Construct (PO)MDP specified by $\mathcal{W}$, A, $\mathcal{T}$, and $\mathcal{R}$ (and $\mathcal{O}$)
13:   Compute policy $\pi$ using $Sol^{pl}$ for the (PO)MDP // Probabilistic planning
14:   **while** $s$ **is not** $term$ **and** $\mathcal{F} \ominus \mathcal{W}^{cplt}$ **do**
15:     Update state $s$ (or belief state $b$) using $\mathcal{F}$
16:     Select action $a$ with $\pi$, and execute $a$
17:     Make an observation to update $\mathcal{F}$ about $\mathcal{V}^{ex}$ // Collect facts $\mathcal{F}$ for detecting exogenous events
18:   **end while**
19: **until** $s$ **is** $term$

---

and checks in Step 14 if it is necessary to break the loop to reconstruct the (PO)MDP to adapt to the exogenous events. It should be noted that not all exogenous events trigger the reconstruction of (PO)MDPs. The necessity is evaluated through a logical operation in Step 14 by

$$\mathcal{F} \ominus \mathcal{W}^{cplt},$$

which reports true when facts $\mathcal{F}$ are consistent with each $w \in \mathcal{W}^{cplt}$, and otherwise reports false. The main loop (Steps 1-19) continues until a terminal state is reached, which identifies the end of a complete trial.

As an example of exogenous events causing inconsistency, consider a robot that plans to avoid the sunlit area (which blinds the range sensors) when it was started. Should clouds appear (an exogenous event) and the previously sunlit area no longer poses a problem to the robot, all possible worlds are rendered inconsistent with the "no cloud" commonsense rule, and the robot reactivates the commonsense reasoner to recompute the MDP state space (and recompute the action policy). Therefore, iCORPP enables the robot's behavior to adapt to the fact of a weather change, without modeling exogenous variable *weather* in its (PO)MDP-based planners.

Next, we look into a few steps of Algorithm 1 providing implementation details, namely Step 4 for logical reasoning, Step 8 for probabilistic reasoning about belief initialization for POMDPs, and Step 11 for probabilistic reasoning to compute the transition function.

***Logical Reasoning with Incomplete Knowledge:*** Since real-world domains are dynamically changing all the time and robots' observations are partial and unreliable, robots frequently need to reason with *incomplete* domain knowledge, and exogenous events.[4] Using ASP, on which P-log is based, our robot can take a set of defaults as input, as a part of $\mathcal{K}^S$, and smoothly revise their values using observed "facts" when available (Step 4), and hence is capable of reasoning with incomplete domain knowledge well. As an example, a robot using an MDP for indoor navigation may have default knowledge: "area $A$ is sunlit in the mornings". A fact of "no sunlight is currently observed in area $A$" can smoothly defeat the default. The set of possible worlds, $\mathcal{W}$, is described by a set of endogenous attributes and their values.

Formally, the logical reasoning step (Step 4) takes as inputs facts $\mathcal{F}$ and logical commonsense knowledge in $\mathcal{K}^S$, and outputs complete possible worlds $\mathcal{W}^{cplt}$. In this step, both endogenous and exogenous variables are reasoned about, and each produced possible world is a complete assignment to both types of variables.

***Probabilistic Reasoning over World States:*** In the case of partially observable domains, we use probabilistic information assignments (or simply probabilistic rules), to compute a probability distribution over the set of possible worlds, i.e., the prior belief of POMDPs.[5] As a result, probabilistic reasoning associates each possible world with a probability $\{\mathcal{W}_0 : pr_0,\ \mathcal{W}_1 : pr_1, \cdots\}$. Step 8 computes a probability, $Pr(w)$, for each possible world whose dimensionality has been reduced in Step 5.

In practice, logical and probabilistic commonsense rules in P-log are processed together using off-the-shelf software packages, i.e., one system that supports the functionalities of both $Sol^{lr}$ and $Sol^{pr}$, and the rules cover both exogenous and endogenous domain attributes ($\mathcal{K}^S$ and $\mathcal{K}^A$). Informally, the steps of logical-probabilistic reasoning about states are to specify the parts of the world that may have effects on the robot working on the current task, i.e., reasoning to "understand" the current world state.

***Probabilistic Reasoning about World Dynamics:*** To represent and reason about state transitions (i.e., world dynamics), we define two identical state spaces using predicates `curr_s` and `next_s` in P-log:

$$\mathtt{curr_s}(\mathtt{V_1}, \cdots, \mathtt{V_n}) \leftarrow \mathtt{v_1} = \mathtt{V_1}, \cdots, \mathtt{v_n} = \mathtt{V_n}.$$
$$\mathtt{next_s}(\mathtt{V_1}, \cdots, \mathtt{V_n}) \leftarrow \mathtt{v'_1} = \mathtt{V_1}, \cdots, \mathtt{v'_n} = \mathtt{V_n}.$$

where `curr_s` and `next_s` specify the current and next states and the `v`'s and `V`'s are endogenous attributes and their variables respectively. If there is at least one endogenous attribute whose value is not directly observable to the robot (Step 7), the corresponding task needs to be modeled as a POMDP (otherwise, an MDP).

We introduce sort `action` and explicitly list a set of $i$ actions, $A$, as a set of *objects* in P-log. Random function `curr_a` maps to one of the actions.

$$\mathtt{action} = \{\mathtt{a_0}, \mathtt{a_1}, \cdots, \mathtt{a_i}\}.$$
$$\mathtt{curr_a} : \mathtt{action}.$$
$$\mathtt{random(curr_a)}.$$

Formally, Step 11 for probabilistic reasoning about transitions takes as inputs a set of possible worlds $\mathcal{W}$ (that is about only $\mathcal{V}^{en}$) and action knowledge $\mathcal{K}^A$, and outputs transition function $T(s, a, s')$, which can be described using a set of pr-atoms in P-log. For instance, the rule below states that the probability of action `A` changing the value of attribute `v` from $\mathtt{V_1}$ to $\mathtt{V_2}$ is 0.9.

$$\mathtt{pr}(\mathtt{v'} = \mathtt{V_2} \mid \mathtt{v} = \mathtt{V_1},\ \mathtt{curr_a} = \mathtt{A}) = 0.9.$$

For MDPs, the values of endogenous attributes are fully observable to the robot, whereas POMDPs need to model a set of observations, $Z$, for estimating the underlying state. We define `obser` as a sort, and `curr_o` as a random function that maps to an observation object `o`.

$$\mathtt{obser} = \{\mathtt{o_0}, \mathtt{o_1}, \cdots, \mathtt{o_j}\}.$$
$$\mathtt{curr_o} : \mathtt{obser}.$$
$$\mathtt{random(curr_o)}.$$

The observation function, $\mathcal{O}$, defines the probability of observing `O` given the current state being `s` and current action being `a`. For instance, the following *pr*-rule states that, if attribute `v`'s current value is `V`, the probability of observing `O` after taking action `A` is 0.8.

$$\mathtt{pr}(\mathtt{curr_o} = \mathtt{O} \mid \mathtt{curr_a} = \mathtt{A},\ \mathtt{v} = \mathtt{V}) = 0.8.$$

The reward function $R$ maps a state-action pair to a numeric value. For instance, the following rule states that taking action `A` given attribute `v`'s value being `V` yields a reward of 10.0.

$$\mathtt{reward}(10.0, \mathtt{A}, \mathtt{V_1}, \cdots, \mathtt{V_n}) \leftarrow \mathtt{curr_a} = \mathtt{A}, \mathtt{curr_s}(\mathtt{V_1}, \cdots, \mathtt{V_n}).$$

Building the reward function of (PO)MDPs requires numerical reasoning, which is not supported by many declarative languages and systems, and hence is not included in the algorithm. In our implementation, we manually encode the reward function using procedural languages.

Next, we describe the design of experiments, and the robot platforms where we implement and evaluate iCORPP.

## 5. Instantiation and Evaluations

In this section, we first describe our experiment design in general, and then introduce two application domains (mobile robot navigation and human-robot dialog). In each of the two domains, we first illustrate the implementation of iCORPP, and then present the experiment results from systematic evaluations.

[4] When we solve an MDP, we simply assume the endogenous attributes are fully observable, and the complete world state is still not fully observable. Robots face a partially observable world in general.

[5] When the current world state is fully observable, there is no need to estimate the current state of the world with observations, and iCORPP uses MDPs for action selections.

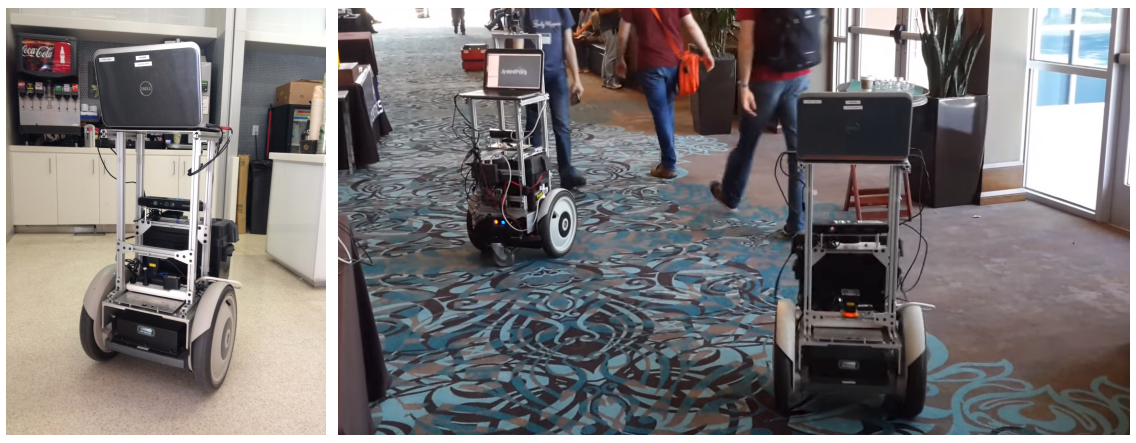

**Figure 4:** The robot platform used in experiments, called BWIBot (Khandelwal et al., 2017). The platform is based on a Segway RMP, and is equipped with sensing capabilities, including laser-based range finding for localization, voice recognition for human-robot dialog, and RGB-depth sensors for human detection and obstacle avoidance. The right is a picture of two BWIBots running at the venue of the Twenty-Ninth AAAI Conference on Artificial Intelligence in Austin, TX.

## 5.1. Experiment Design

iCORPP enables an agent to reason about the current world state and dynamics to construct probabilistic planners (controllers) on the fly, where the reasoning is logical-probabilistic and the planning is based on an MDP or POMDP. Accordingly, experiments were aimed at evaluating the following hypotheses:

(I) Incorporating logical reasoning into probabilistic controllers improves efficiency and accuracy in information gathering;
(II) iCORPP further improves the performance in both accuracy and efficiency by combining logical-probabilistic reasoning and probabilistic controllers;
(III) iCORPP enables fine-tuning of agent behaviors at a level, where a comparable hand-coded controller requires a prohibitively large number of parameters; and
(IV) iCORPP enables agent behaviors that are adaptive to exogenous events, without modeling these exogenous attributes in its controllers.

Baseline algorithms include hand-coded action policies, standard POMDP-based methods, POMDPs with logical reasoning (Zhang et al., 2015a), and the CORPP strategies (Zhang and Stone, 2015).

We used a solver introduced by Zhu (2012) for P-log programs except that reasoning about reward was manually conducted, the APPL solver for POMDPs (Kurniawati et al., 2008), and value iteration for MDPs (Sutton and Barto, 2018).

The iCORPP algorithm has been implemented both in simulation and on real robots. The robot platform used in this study is shown in Figure 4, where the software and hardware were described in a journal article (Khandelwal et al., 2017). The robot is built on top of a Segway Robotic Mobility Platform. It uses a Hokuyo URG-04LX laser rangefinder and a Kinect RGB-D camera for navigation and sensing, and Sphinx-4 (Walker, Lamere, Kwok, Raj, Singh, Gouvea, Wolf and Woelfel, 2004) for speech recognition. The software modules run in Robot Operating System (ROS) (Quigley, Conley, Gerkey, Faust, Foote, Leibs, Wheeler and Ng, 2009). After the proposed approach determines the parameters of the shopping request, it is passed to a hierarchical task planner for creating a sequence of primitive actions that can be directly executed by the robot (Zhang, Yang, Khandelwal and Stone, 2015b).

Experiments in simulation were conducted using Gazebo (Koenig and Howard, 2004), where the environment is shown in Figure 5. In particular, the simulation environment includes a set of human walkers that repeatedly move to arbitrarily-selected navigation goals. The humans can probabilistically block the robot's way.

Next, we evaluate iCORPP using a mobile robot (simulated or physical) that operates in an office environment. Specifically, we use the tasks of *mobile robot navigation* and *spoken dialog system* for illustrating the implementations of iCORPP and system evaluations. The two capabilities together enable the mobile robot to provide a variety of services in human-inhabited environments, such as human guidance, question answering, and object deliveries.

## 5.2. Algorithm Instantiation and Evaluations: Mobile Robot Navigation

Consider a robot navigation problem in a fully-observable 2D grid world shown in Figure 5. The robot can take actions (*North*, *East*, *South*, and *West*) to move toward one of its nearby grid cells, and such actions succeed probabilistically. The area marked with a "sun" is a dangerous area to the robot, because, in the mornings under sunny weather, there is sunlight in areas near east-facing windows that can blind the robot's range-finder sensor, probabilistically causing it to become unrecoverably lost. In this example, the robot's current location should be modeled as an *endogenous* variable, because its value change needs to be modeled in the planning process, i.e., its value needs to be *actively* changed. Current time (morning or not) should be modeled as an *exogenous* variable, meaning that the robot does not need to change its value in the planning process. However, it is indeed necessary to keep an eye on (i.e., to *passively* observe) its value, and adjust the probabilistic planner as needed, e.g., reducing the success rate of navigating though the near-window cell when current time is morning.

### *Algorithm Instantiation*

In the mobile robot navigation task, the state is fully observable, and hence an MDP is used for probabilistic planning. The robot navigates in a domain shown in Figure 5(a). In this domain, people can move and probabilistically block the robot's way, as shown in Figure 5(b). In addition, sunlight can probabilistically blind the robot's laser range-finder, making the robot unrecoverably lost.

Planning is achieved by mapping the domain to a grid, which is defined using sorts `row` and `col`, and predicates `belowof` and `leftof`.

$$\texttt{row} = \{\texttt{rw0}, \texttt{rw1}, \cdots, \texttt{rw4}\}.$$
$$\texttt{col} = \{\texttt{cl0}, \texttt{cl1}, \cdots, \texttt{cl5}\}.$$

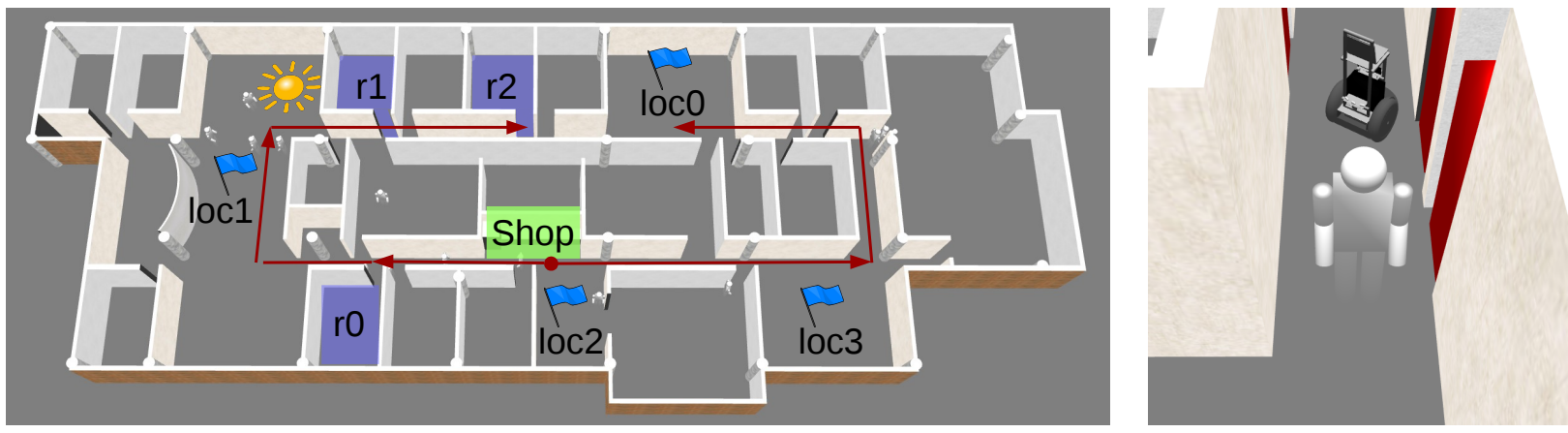

(a) (b)

**Figure 5:** (a) Simulation environment used in experiments, where the red arrows indicate the delivery routes from the shop to individual rooms; and (b) A human walker blocking the way of the robot.

$$\texttt{leftof(cl0, cl1).} \cdots \texttt{leftof(cl4, cl5).}$$
$$\texttt{belowof(rw1, rw0).} \cdots \texttt{belowof(rw4, rw3).}$$

We then introduce predicates `near_row` and `near_col` used for specifying if two grid cells are next to each other, where R's (C's) are variables of row (column).

$$\texttt{near_row}(\texttt{RW}_1, \texttt{RW}_2) \leftarrow \texttt{belowof}(\texttt{RW}_1, \texttt{RW}_2).$$
$$\texttt{near_row}(\texttt{RW}_1, \texttt{RW}_2) \leftarrow \texttt{near_row}(\texttt{RW}_2, \texttt{RW}_1).$$
$$\texttt{near_col}(\texttt{CL}_1, \texttt{CL}_2) \leftarrow \texttt{leftof}(\texttt{CL}_1, \texttt{CL}_2).$$
$$\texttt{near_col}(\texttt{CL}_1, \texttt{CL}_2) \leftarrow \texttt{near_col}(\texttt{CL}_2, \texttt{CL}_1).$$

To model the non-deterministic action outcomes, we define random functions `curr_row` and `next_row` that map to the current and next rows, and `curr_col` and `next_col` that map to the current and next columns. For instance, the first of the following two random rules states that, given the robot's current row is RW, it will be in row R_ in the next step, where R_ and RW are adjacent rows, i.e., `near_row(R_, RW)` is true.

$$\texttt{random}(\texttt{next_row} : \{\texttt{R_} : \texttt{near_row}(\texttt{R_}, \texttt{RW})\})$$
$$\leftarrow \texttt{curr_row} = \texttt{RW}.$$
$$\texttt{random}(\texttt{next_col} : \{\texttt{C_} : \texttt{near_col}(\texttt{C_}, \texttt{CL})\})$$
$$\leftarrow \texttt{curr_col} = \texttt{CL}.$$

We use predicates `near_window` and `sunlit` to define the cells that are near to a window and the sunlit cells. The P-log rule below is a default stating that: in the mornings, a cell near a window is believed to be under sunlight, unless the statement is defeated elsewhere.

$$\texttt{sunlit}(\texttt{RW}, \texttt{CL}) \leftarrow \texttt{near_window}(\texttt{RW}, \texttt{CL}),$$
$$\texttt{not}\ \neg\texttt{sunlit}(\texttt{RW}, \texttt{CL}),$$
$$\texttt{curr_time} = \texttt{morning}.$$

The above-mentioned rules are logical state knowledge in $\mathcal{K}^S$ as defined in Algorithm 1. While navigating in areas under sunlight, there is a large probability of becoming lost (0.9), which deterministically leads to the end of an episode.

$$\texttt{pr}(\texttt{next_term} = \texttt{true} \mid \texttt{curr_row} = \texttt{RW},$$
$$\texttt{curr_col} = \texttt{CL},$$
$$\texttt{sunlit}(\texttt{RW}, \texttt{CL})) = 0.9.$$

$$\texttt{pr}(\texttt{next_term} = \texttt{true} \mid \texttt{curr_term} = \texttt{true}) = 1.0.$$

In this domain, `curr_row`, `curr_col`, and `curr_term` (and their "`next_`" counterparts) are endogenous variables $\mathcal{V}^{en}$, meaning that the robot actively changes their values, and all the other variables are exogenous.

The robot can take actions to move to a grid cell next to its current one: $\texttt{action} = \{\texttt{left}, \texttt{right}, \texttt{up}, \texttt{down}\}$. Action knowledge is defined in $\mathcal{K}^A$. For instance, given action `up`, the probability of successfully moving to the above grid cell is 0.9, given no obstacle is in the above cell.

$$\texttt{pr}(\texttt{next_row}{=}\ \texttt{RW}_2 \mid \texttt{curr_row} = \texttt{RW}_1,\ \texttt{curr_col} = \texttt{CL}_1,$$
$$\texttt{belowof}(\texttt{RW}_1, \texttt{RW}_2),\ \neg\texttt{sunlit}(\texttt{RW}_2, \texttt{CL}_1),$$
$$\neg\texttt{blocked}(\texttt{RW}_2, \texttt{CL}_1),\ \texttt{curr_a} = \texttt{up}) = 0.9.$$

Finally, the current state is specified by endogenous attributes `curr_row`, `curr_col`, and `curr_term`:

$$\texttt{curr_state}(\texttt{RW}, \texttt{CL}, \texttt{TM}) \leftarrow \texttt{curr_row} = \texttt{RW},$$
$$\texttt{curr_col} = \texttt{CL},$$
$$\texttt{curr_term} = \texttt{TM}.$$

The goal of visiting room (r0, c3) can be defined as below, where successfully arriving at the goal room produces a reward of 50 and an early termination causes a reward of $-100.0$ (i.e., a penalty of 100).

$$\texttt{pr}(\texttt{next_term} = \texttt{true} \mid \texttt{curr_row} = \texttt{r0},$$
$$\texttt{curr_col} = \texttt{c3}) = 1.0.$$
$$\texttt{reward}(50.0, \texttt{A}, \texttt{r0}, \texttt{c3}, \texttt{true})$$
$$\leftarrow \texttt{curr_state}(\texttt{r0}, \texttt{c3}, \texttt{true}).$$
$$\texttt{reward}(-100.0, \texttt{A}, \texttt{RW}, \texttt{CL}, \texttt{true})$$
$$\leftarrow \texttt{curr_state}(\texttt{RW}, \texttt{CL}, \texttt{true}), \texttt{RW} <> \texttt{r0}.$$
$$\texttt{reward}(-100.0, \texttt{A}, \texttt{RW}, \texttt{CL}, \texttt{true})$$
$$\leftarrow \texttt{curr_state}(\texttt{RW}, \texttt{CL}, \texttt{true}), \texttt{CL} <> \texttt{c3}.$$

### *Evaluation*

Since the robot navigation domain is highly dynamic with human walkers, we focus on evaluating the hypothesis of the robot being able to adapt its behaviors to exogenous

**Table 1**
Average time (with standard deviations) consumed in navigating between location pairs when awalker moves near the door of room $r1$

| Navigation | Positions | | | | | |
|---|---|---|---|---|---|---|
| Start | loc0 | loc0 | loc0 | loc1 | loc1 | loc2 |
| Goal | loc1 | loc2 | loc3 | loc2 | loc3 | loc3 |
| Strategies | Average navigation time (second) | | | | | |
| Stationary policy | 193.15 (38.64) | 252.47 (29.51) | 82.40 (1.38) | 59.65 (2.44) | 94.81 (1.49) | 59.97 (1.03) |
| iCORPP | 151.01 (1.89) | 115.82 (1.65) | 81.78 (1.74) | 60.48 (1.00) | 94.86 (1.21) | 60.06 (1.22) |

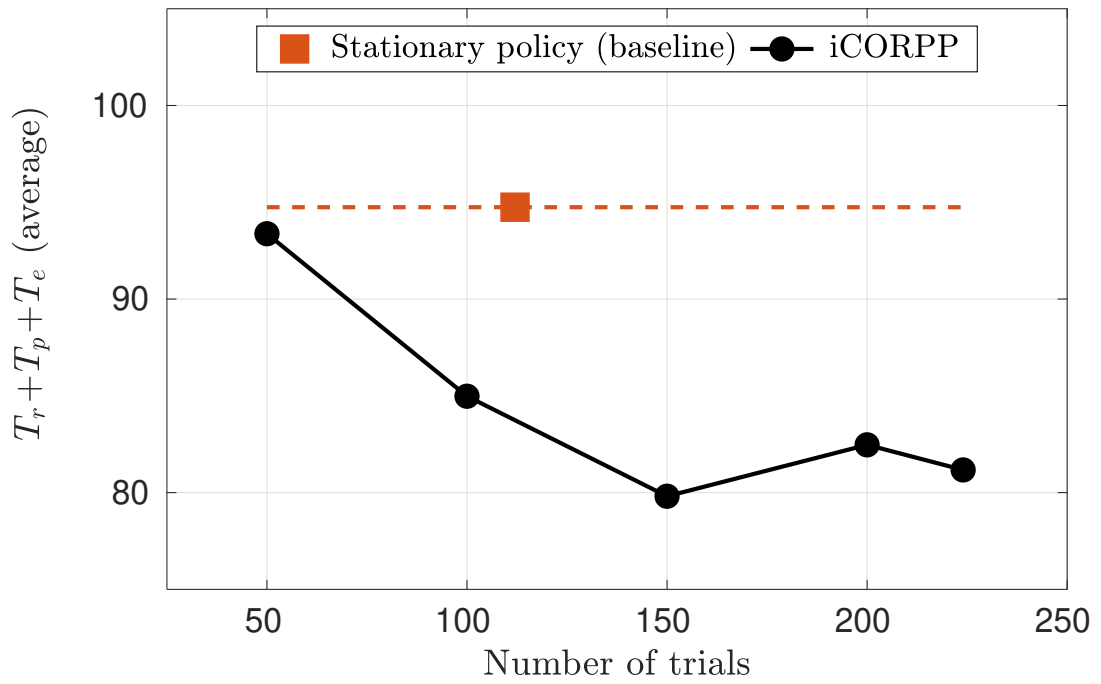

**Figure 6:** iCORPP enables the robot to adapt to exogenous domain changes (the walker's position). Results are processed in batches (each has 50 trials, when available).

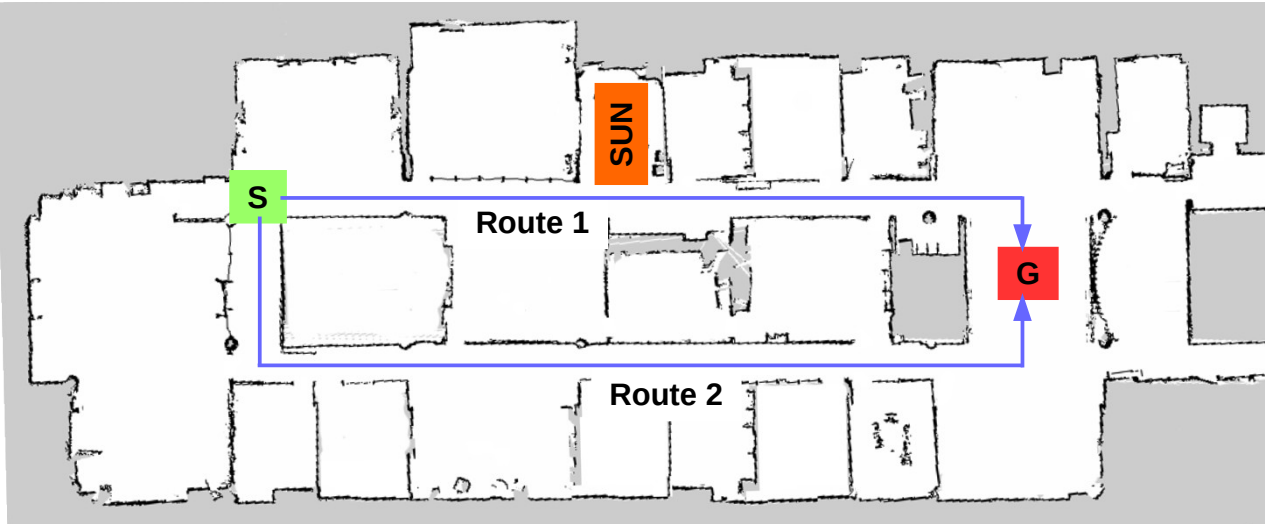

**Figure 7:** iCORPP enables the robot to select "Route 1", successfully avoiding the "sunlight" area along "Route 2".

events (human positions in this case), i.e., Hypothesis IV as listed in Section 5.1.

The testing environment and the robot are shown in Figure 5(a) and 5(b). We limit the number of random walkers that affect robot navigation actions to be 1 and its speed to be one fifth of the robot's. This setting ensures no human-robot collisions given the robot's intention and capability of obstacle avoidance. A goal room is randomly selected from the four flag rooms. Reasoning happens only after the current episode is terminated (goal room is reached). The walker's position is the only exogenous domain change (by temporarily setting the time to be "evening"). We cached policies for both the baseline that uses stationary policies (four policies corresponding to four goal rooms) and iCORPP (56 policies).

Table 1 shows the robot's traveling time given start-goal pairs: once the robot arrives at its current goal, the next one is randomly selected. The walker moves slowly near the door of room $r1$. Without adaptive planning developed in this work, the robot follows the "optimal" path and keeps trying to bypass the walker for a fixed length of time. If the low-level motion planner does not find a way to bypass the walker within the time, the robot will take the other way to navigate to the other side of the walker and continues executing the "optimal" plan generated by the outdated model. We can see when the robot navigates between $loc0$ and $loc2$, iCORPP reduces the traveling time from about 250 seconds to about 110 seconds, producing a significant improvement. We do not see a significant difference for position pairs other than "0-1" and "0-2", because the walking human is constrained to be near the door of room $r1$.

Results over 8.5 hours of experiments (wall-clock time) are shown in Figure 6: 224 trials using iCORPP and 112 trials using the baseline of stationary policies. Without caching, we find the time consumed by iCORPP (over 54 trials) is distributed over P-log reasoning ($T_r$, 28%), MDP planning ($T_p$, <1%), and execution ($T_e$, 72%). Compared to the baseline, iCORPP enables the robot to spend much less time in execution ($T_e$) in all phases. At the beginning phase, iCORPP requires more reasoning time for dynamically constructing MDPs, which together with the less execution time makes the overall time comparable to the baseline (left ends of Figure 6). Eventually, the low execution time ($T_e$) dominates the long-term performance (right ends of Figure 6), supporting that iCORPP enables the robot to adapt to exogenous domain changes, whereas stationary policies can not.

Figure 7 shows the office environment where real-robot experiments were conducted. It includes ten offices, two meeting rooms, and three research labs. The occupancy-grid map of the environment was generated using a simultaneous localization and mapping (SLAM) algorithm. The "SUN" area is an area that is subject to strong sunlight in the mornings given the current weather being sunny. The *default reasoning* capability of P-log supports that a fact of "under sunlight" (or not) can defeat the default belief about sunlight. Such sunlight can blind the robot's laser range-finder, and makes the robot unrecoverably lost. Therefore, the robot needs to reason about the knowledge of current time and weather to dynamically construct its MDP-based probabilistic transition system, including the success rate of navigating through the "SUN" area given the current condition. Figure 7 also shows two routes in a demonstration trial where the robot needs to navigate from its start point ("S" in the green box) to the goal ("G" in the red box).

To test the robot's behavior adapting to sunlight change, we left the robot two routes that lead to the goal. For instance, Route 1 is shorter, but it goes through the area that is currently under sunlight. Figure 8 shows screenshots of two trials in which the baseline of stationary policies and iCORPP were used respectively. iCORPP enables the robot to select the safer route (Route 2), even though it is longer. the baseline strategy cannot adapt to the exogenous change of current time being morning and current weather being sunny, letting the robot still believe the shorter path is safe. In experiments, we directly encode such exogenous knowledge to the robot. Demo videos of simulated and real-robot trials are available at: `https://youtu.be/QvuWLuGjsOY`

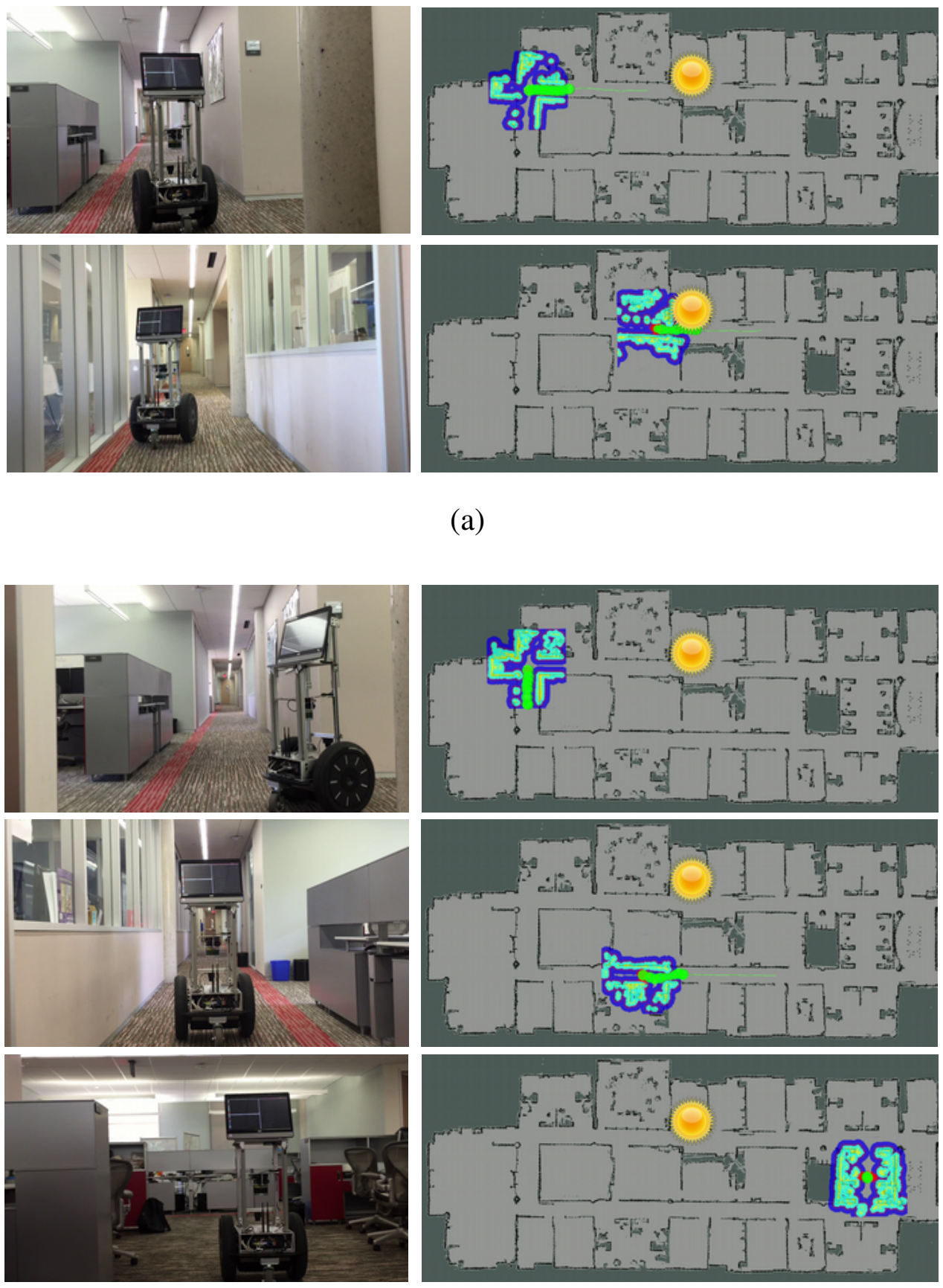

(a)

(b)

**Figure 8:** Screen shots of two illustrative trials. (a) Using the baseline approach (CORPP), the robot chose Route 1 that is dangerous but shorter, causing the robot to become unrecoverably lost in the sunlit area; (b) By reasoning about current time (morning) and weather (sunny), iCORPP successfully helps the robot take Route 2 to avoid being trapped in the "sunlight" area, even though the route is longer.

## 5.3. Algorithm Instantiation and Evaluations: Spoken Dialog Systems

In this section, we present an instantiation of iCORPP in a second evaluation domain, namely spoken dialog system (SDS). The main difference from mobile robot navigation (Section 5.2) is that the current state of the world in SDSs (i.e., the dialog state) is partially observable to the agent. As a result, dialog agents use observations to maintain a belief over possible dialog states. To account for the partial observability, we use POMDPs for probabilistic planning in spoken dialog systems. Accordingly, Step 8 in Algorithm 1, where iCORPP reasons to compute a prior belief distribution for state estimation, is activated in this instantiation. Next, we briefly summarize the key components of complete SDSs, and then present the instantiation details.

A spoken dialog system enables an agent to interact with a human using speech, and typically has three key components: spoken language understanding (SLU), dialog management (DM), and natural language generation (NLG). SLU takes speech from humans and provides semantic representations to a dialog manager; DM uses the semantic representations to update its internal state $s$ and uses a policy $\pi$ to determine the next language action; and NLG converts the action back into speech. Despite significant improvements in speech recognition over the past decades, e.g., the work of Graves, Mohamed and Hinton (2013), it is still a challenge to reliably understand spoken language, especially in robotic domains. POMDPs have been used in dialog management (spoken and text-based) to account for the uncertainties from SLU by maintaining a *distribution* (as a POMDP belief state) over all possible user meanings. Solving a POMDP problem generates a policy that maps the current belief state to an optimal action (an utterance by the system). Young, Gasic, Thomson and Williams (2013) reviewed existing POMDP-based spoken dialog systems.

### *Algorithm Instantiation*

In a campus environment, the mobile robot can help buy an item for a person and deliver to a room, so a shopping request is in the form of $\langle$*item, room, person*$\rangle$. The ontology of items is shown in Figure 9. The distances between rooms are shown in Figure 5(a). A person can be either a *professor* or a *student*. Registered students are authorized to use the robot and professors are not unless they paid. The robot can get access to a database to query about registration and payment information, but the database may be incomplete. The robot can initiate spoken dialog to gather information for understanding shopping requests and take a delivery action when it becomes confident in the estimation. This task is challenging for the robot because of its imperfect speech recognition ability. The goal is to identify shopping requests, e.g., $\langle$*coffee, office*1*, alice*$\rangle$, efficiently and accurately. The original shopping request identification problem, which requires a spoken dialog system, was presented in our previous work (Zhang and Stone, 2015).

Unlike the navigation task, the current dialog state is partially observable to the robot, and has to be estimated using observations via POMDPs. We use this task to illustrate constructions of POMDP-based controllers on the fly, and evaluate how iCORPP enables the robot to adapt to exogenous domain changes (e.g., missing items in the ontology) and fine-tune its behaviors.

***Logical Reasoning about States:*** This domain has the following *sorts*, $\Theta$, and each sort has a set of objects.

$$\texttt{time} = \{\texttt{morning}, \texttt{noon}, \cdots\}. \quad \texttt{room} = \{\texttt{r0}, \texttt{r1}, \cdots, \texttt{shop}, \cdots\}.$$

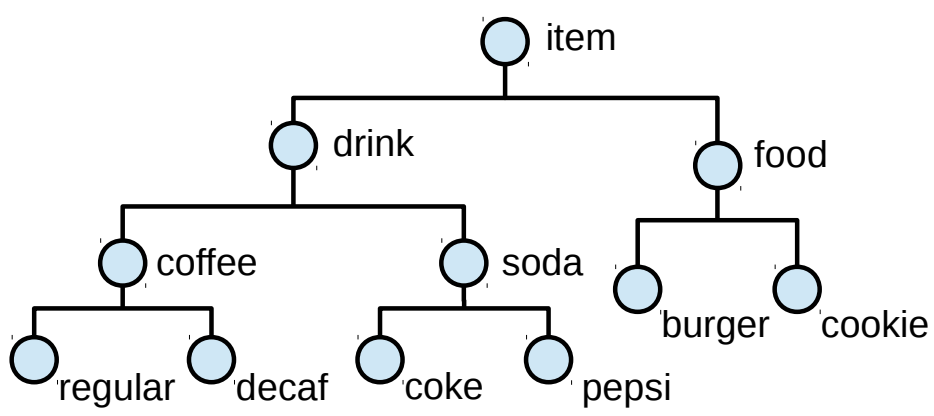

**Figure 9:** An ontology of available items used in the "shopping" task. This ontology is used for two purposes: 1) the evaluation of iCORPP generating robot behaviors that adapt to dynamic changes (e.g., missing items in this ontology); and 2) the evaluation of iCORPP generating fine-tuned behaviors.

$$\texttt{person} = \{\texttt{alice}, \texttt{bob}, \cdots\}. \quad \texttt{item} = \{\texttt{regular}, \texttt{decaf}, \cdots\}.$$
$$\texttt{class} = \{\texttt{item}, \texttt{drink}, \texttt{food}, \texttt{coffee}, \texttt{soda}\}.$$

We then define predicate set $\mathcal{P}$ : $\{\texttt{request}, \texttt{subcls}\}$, where $\texttt{request(I,R,P)}$ specifies a shopping request of delivering item $\texttt{I}$ to room $\texttt{R}$ for person $\texttt{P}$, and $\texttt{subcls}(\texttt{C}_1, \texttt{C}_2)$ claims class $\texttt{C}_1$ to be a subclass of class $\texttt{C}_2$. Figure 9 shows the categorical tree that can be represented using rules:

$$\texttt{subcls}(\texttt{C}_1, \texttt{C}_3) \leftarrow \texttt{subcls}(\texttt{C}_1, \texttt{C}_2),\ \texttt{subcls}(\texttt{C}_2, \texttt{C}_3).$$
$$\texttt{is}(\texttt{I}, \texttt{C}_1) \leftarrow \texttt{is}(\texttt{I}, \texttt{C}_2),\ \texttt{subcls}(\texttt{C}_2, \texttt{C}_1).$$

and other predicates include:

$$\texttt{place(P,R). prof(P). student(P). registered(P).}$$
$$\texttt{authorized(P). paid(P). task(I,R,P).}$$

where $\texttt{place(P,R)}$ represents person $\texttt{P}$'s working room is $\texttt{R}$, $\texttt{authorized(P)}$ states P is authorized to place orders, and a ground of $\texttt{task(I,R,P)}$ specifies a shopping request.

The following two logical reasoning rules state that professors who have paid and students who have registered are authorized to place orders.

$$\texttt{authorized(P)} \leftarrow \texttt{paid(P)},\ \texttt{prof(P)}.$$
$$\texttt{authorized(P)} \leftarrow \texttt{registered(P)},\ \texttt{student(P)}.$$

Since the database can be incomplete about the registration and payment information, we need default knowledge to reason about unspecified variables. For instance, if it is unknown that a professor has paid, we believe the professor has not; if it is unknown that a student has registered, we believe the student has not.

$$\neg\texttt{paid(P)} \leftarrow \texttt{not paid(P)},\ \texttt{prof(P)}.$$
$$\neg\texttt{registered(P)} \leftarrow \texttt{not registered(P)},\ \texttt{student(P)}.$$

ASP is strong in default reasoning in that it allows prioritized defaults and exceptions at different levels (Gelfond and Kahl, 2014). There is the Closed World Assumption (CWA) in logical reasoning for some predicates, e.g., the below rule guarantees that the value of attribute $\texttt{authorized(P)}$ must be either $\texttt{true}$ or $\texttt{false}$ (cannot be unknown):

$$\neg\texttt{authorized(P)} \leftarrow \texttt{not authorized(P)}.$$

To identify a shopping request, the robot always starts with collecting all available facts, e.g.,

$$\texttt{prof(alice). prof(bob). prof(carol). student(dan).}$$
$$\texttt{student(erin). place(alice,office1).}$$
$$\texttt{place(bob,office2). place(erin,lab).}$$

If the robot also observes facts of $\texttt{paid(alice)}$, $\texttt{paid(bob)}$ and $\texttt{registered(dan)}$, reasoning with the above defaults and rules will imply that $\texttt{alice}$, $\texttt{bob}$ and $\texttt{dan}$ are authorized to place orders. Thus, logical reasoning produces a set of possible worlds by reasoning with the rules, facts and defaults.

***Probabilistic Reasoning about States:*** A set of random functions describes the possible values of random variables: $\texttt{curr_time}$, $\texttt{req_item(P)}$, $\texttt{req_room(P)}$, and $\texttt{req_person}$. E.g., the two rules below state that if the delivery is for person $\texttt{P}$, the value of $\texttt{req_item}$ is randomly selected from the range of $\texttt{item}$, unless fixed elsewhere:

$$\texttt{random(req_item(P))}.\ \ \texttt{req_item} : \texttt{person} \rightarrow \texttt{item}.$$

We can then use a *pr-atom* to specify a probability. For instance, the rule below states that the probability of delivering $\texttt{coffee}$ in the morning is 0.8.

$$\texttt{pr(req_item(P)} = \texttt{coffee} | \texttt{curr_time} = \texttt{morning}) = 0.8.$$

Such random selection rules and pr-atoms allow us to represent and reason with commonsense with probabilities. In this domain, the "req_" variables are endogenous, where the agent actively observes their values. Finally, a shopping request is specified as follows:

$$\texttt{task(I,R,P)} \leftarrow \texttt{req_item(P)} = \texttt{I},\ \texttt{req_room(P)} = \texttt{R},$$
$$\texttt{req_person} = \texttt{P},\ \texttt{authorized(P)}.$$

The P-log reasoner takes queries from the POMDP-based planner and returns the joint probability. For instance, if it is known that Bob, as a professor, has paid and the current time is morning, a query for calculating the probability of $\langle\texttt{sandwich}, \texttt{office1}, \texttt{alice}\rangle$ is of the form:

$$?\{\texttt{task(sandwich,office1,alice)}\} \mid \texttt{do(paid(bob))},$$
$$\texttt{obs(curr_time} = \texttt{morning}).$$

The fact of $\texttt{bob}$ having paid increases the uncertainty in estimating the value of $\texttt{req_person}$ by bringing additional possible worlds that include $\texttt{req_person} = \texttt{bob}$.

***Reasoning about Actions:*** The action set is explicitly enumerated as below.

$$\begin{aligned} \texttt{action} = \{&\texttt{ask_i}, \texttt{ask_r}, \texttt{ask_p}, \texttt{conf_i0}, \texttt{conf_i1}, \cdots, \\ &\texttt{conf_r0}, \texttt{conf_r1}, \cdots, \texttt{conf_p0}, \texttt{conf_p1}, \cdots, \\ &\texttt{del_i0_r0_p0}, \texttt{del_i0_r0_p1}, \cdots\} \end{aligned}$$

where, `ask_`'s are general questions (e.g., `ask_r` corresponds to "which room to deliver?"), `conf_`'s are confirming questions (e.g., `conf_r0` corresponds to "is this delivery to `room0`?"), and `del_`'s are actions of deliveries.

For delivery actions, the reward function $\mathcal{R}$ maps a state-action pair to a real number, and is defined as:

$$\mathcal{R}(a^{del}, s) = \begin{cases} R^{+}, & \textbf{if } a_i \odot s_i \textbf{ and } a_p \odot s_p \textbf{ and } a_r \odot s_r \\ \big(1 - \lambda_i(a_i, s_i) \cdot \lambda_p(a_p, s_p) \cdot \lambda_r(a_r, s_r)\big) R^{-}, & \textbf{otherwise} \end{cases}$$

where operator $\odot$ returns true if the action on the left matches the state on the right in the given dimension (subscript). $\lambda$ in the range of (0, 1] measures the closeness between actual delivery (action) and underlying request (state) in item, person, and room, respectively. $R^{+}$ and $R^{-}$ are the reward and penalty that a robot can get in extreme cases (completely correct or completely incorrect deliveries).

We compute the closeness of two items, $\lambda(I_1, I_2)$ by post-processing the resulting answer set. Specifically, the heuristic closeness function of two items is defined as:

$$\lambda_i(I_1, I_2) = 1 - \frac{max\big(dep(LCA, I_1), dep(LCA, I_2)\big) - 1}{max\big(dep(root, I_1), dep(root, I_2)\big)} \tag{2}$$

where $LCA$ is the lowest common ancestor of $I_1$ and $I_2$ and *dep(C,I)* is the number of nodes (inclusive) between $C$ and $I$.

Informally, the closeness of room $R_1$ to room $R_2$ is inversely proportional to the effort needed to recover from a delivery to $R_1$ given the request being to $R_2$. In Figure 5(a), for instance, a wrong delivery to $r0$ given the request being to $r1$ requires the robot to go back to shop, learn the delivery room being $r1$, and then move to room $r1$. Therefore, the *asymmetric* room closeness function is defined as below:

$$\lambda_r(R_1, R_2) = \frac{dis(shop, R_2)}{2 \cdot dis(shop, R_1) + dis(shop, R_2)} \tag{3}$$

We simply set $\lambda_p$ to 1. The costs of question-asking actions are stationary: $\mathcal{R}(a^{ask}, s)$=-1, and $\mathcal{R}(a^{conf}, s)$=-2.

***Probabilistic Planning with POMDPs:*** A POMDP needs to model all partially observable attributes relevant to the task at hand. In the shopping request identification problem, an underlying state is composed of an item, a room and a person. The robot can ask polar questions such as "*Is this delivery for Alice?*", and wh-questions such as "*Who is this delivery for?*". The robot expects observations of "yes" or "no" after polar questions and an element from the sets of items, rooms, or persons after wh-questions. Once the robot becomes confident in the request estimation, it can take a delivery action that deterministically leads to a terminating state. Each delivery action specifies a shopping task.

- $\mathcal{S} : \mathcal{S}_i \times \mathcal{S}_r \times \mathcal{S}_p \cup$ *term* is the state set. It includes a Cartesian product of the set of items $\mathcal{S}_i$, the set of rooms $\mathcal{S}_r$, and the set of persons $\mathcal{S}_p$, and a terminal state *term*.
- $\mathcal{A} : \mathcal{A}_w \cup \mathcal{A}_p \cup \mathcal{A}_d$ is the action set. $\mathcal{A}_w = \{a^i_w, a^r_w, a^p_w\}$ includes actions of asking wh-questions. $\mathcal{A}_p = \mathcal{A}^i_p \cup \mathcal{A}^r_p \cup \mathcal{A}^p_p$ includes actions of asking polar questions, where $\mathcal{A}^i_p$, $\mathcal{A}^r_p$ and $\mathcal{A}^p_p$ are the sets of actions of asking about items, rooms and persons respectively. $\mathcal{A}_d$ includes the set of delivery actions. For $a \in \mathcal{A}_d$, we use $s \odot a$ to represent that the delivery of $a$ matches the underlying state $s$ (i.e., a correct delivery) and use $s \oslash a$ otherwise.
- $T : \mathcal{S} \times \mathcal{A} \times \mathcal{S} \rightarrow [0, 1]$ is the state transition function. Action $a \in \mathcal{A}_w \cup \mathcal{A}_p$ does not change the state and action $a \in \mathcal{A}_d$ results in the terminal state *term* deterministically.
- $Z : Z_i \cup Z_r \cup Z_p \cup \{z^{+}, z^{-}\}$ is the set of observations, where $Z_i$, $Z_r$ and $Z_p$ include observations of action *item*, *room* and *person* respectively. $z^{+}$ and $z^{-}$ are the positive and negative observations after polar questions.
- $O : \mathcal{S} \times \mathcal{A} \times Z \rightarrow [0, 1]$ is the observation function. The probabilities of $O$ are empirically hand-coded, e.g., $z^{+}$ and $z^{-}$ are more reliable than other observations. Learning the probabilities is beyond the scope of this article.
- $R : \mathcal{S} \times \mathcal{A} \rightarrow \mathbb{R}$ is the reward function. In our case:

$$R(s, a) = \begin{cases} -r_p, & \textbf{if } s \in \mathcal{S},\ a \in \mathcal{A}_p \\ -r_w, & \textbf{if } s \in \mathcal{S},\ a \in \mathcal{A}_w \\ -r^{-}_d, & \textbf{if } s \in \mathcal{S},\ a \in \mathcal{A}_d,\ s \oslash a \\ r^{+}_d, & \textbf{if } s \in \mathcal{S},\ a \in \mathcal{A}_d,\ s \odot a \end{cases} \tag{4}$$

where we use $r_w$ and $r_p$ to specify the costs of asking wh- and polar questions. $r^{-}_d$ is a big cost for an incorrect delivery and $r^{+}_d$ is a big reward for a correct one. Unless otherwise specified, $r_w = 1$, $r_p = 2$, $r^{-}_d = 100$, and $r^{+}_d = 50$.

Consider an example where $\mathcal{S}_i = \{$*coffee, sandwich*$\}$, $\mathcal{S}_r = \{$*lab*$\}$, and $\mathcal{S}_p = \{$*alice,bob*$\}$. The state set will be specified as: $\mathcal{S} = \{$*coffee_lab_alice*$, \ldots,$ *term*$\}$ with totally five states, where each state corresponds to a possible world specified by a set of literals (a `task` in our case), and *term* corresponds to the possible world with no `task`.

The corresponding action set $\mathcal{A}$ will have 12 actions with $|\mathcal{A}_w| = 3$, $|\mathcal{A}_p| = 5$, and $|\mathcal{A}_d| = 4$. Observation set $Z$ will be of size $|Z| = 7$ including $z^+$ and $z^-$ for polar questions.

Given a POMDP, we calculate a policy using state-of-the-art POMDP solvers, e.g., APPL (Kurniawati et al., 2008). The policy maps a POMDP belief to an action toward maximizing the long-term rewards. Specifically, the policy enables the robot to take a delivery action only if it is confident enough about the shopping request that the cost of asking additional questions is not worth the expected increase in confidence. The policy also decides *what*, to *whom* and *where* to deliver. There are attributes that contribute to calculating the POMDP priors but are irrelevant to the optimal policy given the prior. The reasoning components shield such attributes, e.g., `curr_time`, from the POMDPs.

iCORPP enables the dialog manager (for identifying shopping requests) to combine commonsense reasoning with probabilistic planning. For instance, reasoning with the commonsense rule of "people usually buy coffee in the morning" and the fact of current time being morning, our robot prefers "Would you want to buy coffee?" to a wh-question such as "What item do you want?" in initiating a conversation. At the same time, the POMDP-based planner ensures the robustness to speech recognition errors.

### *Experiments using Spoken Dialog Systems*

We first define three straightforward policies that gather information in a pre-defined way. They serve as comparison points representing easy-to-define hand-coded policies.

- **Defined-1** allows the robot to take actions from $\mathcal{A}_w$;
- **Defined-2** allows actions from $\mathcal{A}_p$; and
- **Defined-3** allows actions from $\mathcal{A}_w \cup \mathcal{A}_p$.

We further define a *round* as taking all allowed actions, once for each. In the end of a trial, the robot finds the shopping request (corresponding to state $s$) that it is most certain about and then takes action $a \in \mathcal{A}_d$ to maximize the probability of $s \odot a$. The robot does not necessarily have full and/or accurate probabilistic commonsense knowledge. We distinguish the probabilistic knowledge provided to the robot based on its availability and accuracy. Each data point in the figures in this section is the average of at least 10,000 simulated trials.

- **All**: the robot can get access to the knowledge described in Section 5.3 in a complete and accurate way;
- **Limited**: the accessibility to the knowledge is the same as "All" except that current time is hidden from the robot.
- **Inaccurate**: the accessibility to the knowledge is the same as "All" except that the value of current time is always wrong.

We compared the POMDP-based probabilistic planner against the three defined information gathering policies. All start with uniform $\alpha$ meaning that *all* worlds are equally probable. The defined policies can gather information by taking multiple rounds of actions. The results are shown as the *hollow markers* in Figure 10. The POMDP-based controller enables the delivery requests to be correctly identified in more than 90% of the trials with costs of about 14.3 units on average (black hollow square) with the imperfect sensing ability. In contrast, the defined policies need more cost (e.g., about 44 units for Defined-2) to achieve comparable accuracy (red hollow circle). Therefore, POMDP-based planning enables efficient and accurate information gathering and behavior in identifying delivery requests.

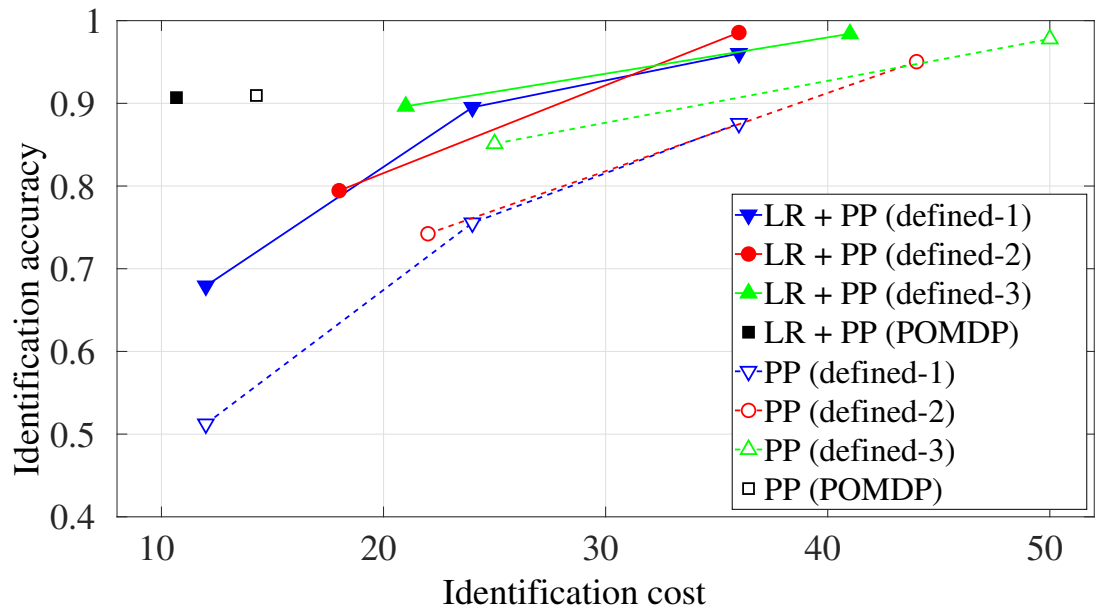

**Figure 10:** POMDP-based probabilistic planner performs better than the defined baseline policies in efficiency and accuracy; and combining PP with logical reasoner further improves the performance (Hypothesis-I).

In the following figures, we use LR, PR, and PP to represent logical reasoning, probabilistic reasoning, and probabilistic planning respectively.

To evaluate Hypothesis-I (integrated POMDP-based probabilistic planning with logical reasoning), the POMDP-based controller and the three defined policies are next combined with logical reasoning. Here, logical reasoning is realized using logical rules, defaults, and facts, and results are shown as the *solid markers* in Figure 10. Without probabilistic knowledge, we can only assume all logically possible worlds to be equally probable in calculating the prior $\alpha$. We can see the combination of logical reasoning and POMDP-based planning performs better than the combination of LR and the three defined planning policies—see the *solid markers*. Furthermore, comparing to the corresponding hollow markers, we can see adding logical knowledge improves the performance of both probabilistic planning and the defined policies. Specifically, logical reasoning enables the POMDP-based planner to reduce the average cost to about 10.5 units without hurting the accuracy. Logical reasoning reduces the number of possible worlds (from 40 to 24 in our case), which enables POMDP solvers to calculate more accurate action policies in reasonable time (an hour in our case) and reduces the uncertainty in state estimation.

Next, we focus on evaluating Hypothesis-II: integrated logical-probabilistic reasoning and probabilistic planning (referred to as LR+PR+PP) performs better than "planning only" (referred to as PP) and "integrated logical reasoning and probabilistic planning" (referred to as LR+PP). We provide the probabilistic commonsense knowledge to the robot at different completeness and accuracy levels—learning the probabilities is beyond the scope of this article. Experimental results are shown in Figure 11.[6] Each set of experiments has three data points because we assigned different penalties to incorrect identifications in PP ($r_d^-$ equals 10, 60 and 100). Generally, a larger penalty requires the robot to ask more questions before taking a delivery action. POMDP-based probabilistic planning without commonsense reasoning (blue rightward triangle) produced the worst results. Combining logical reasoning with probabilistic planning (magenta leftward triangle) improves the performance by reducing the number of possible worlds. Adding *inaccurate* probabilistic commonsense (green upward triangle) hurts the accuracy significantly when the penalty of incorrect identifications is small. Reasoning with *limited* probabilistic commonsense requires much less cost and results in higher (or at least similar) accuracy on average, compared to planning without probabilistic reasoning. Finally, the proposed algorithm, iCORPP, produced the best performance in both efficiency and accuracy. We also find that the POMDP-based PP enables the robot to recover from inaccurate knowledge by actively gathering more information—compare the right ends of the "*limited*" and "*inaccurate*" curves.

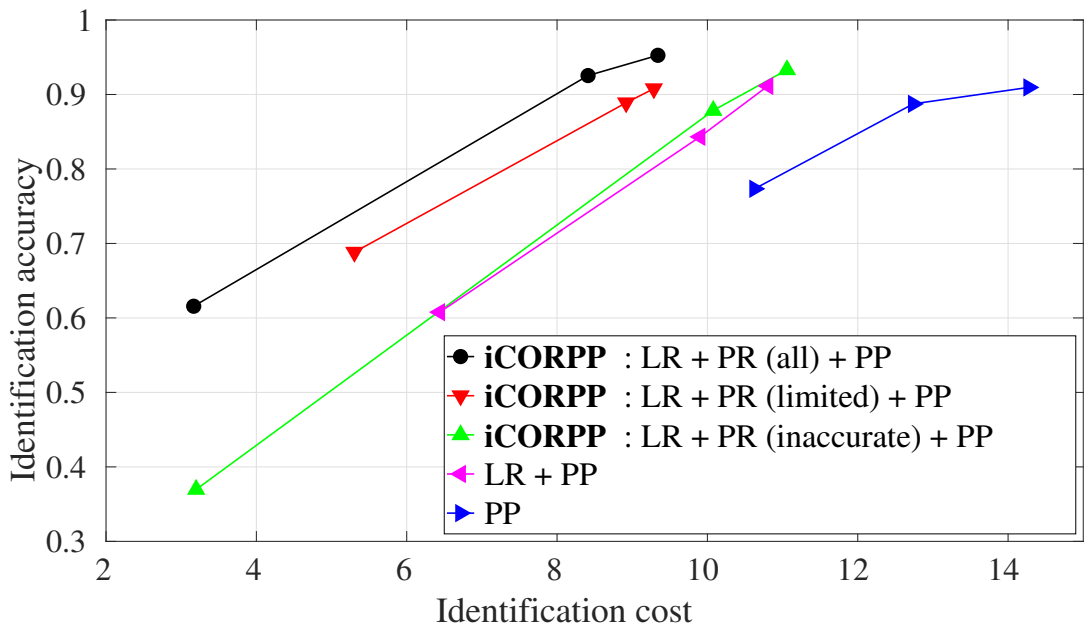

**Figure 11:** iCORPP performs better than the other approaches in both efficiency and accuracy (Hypothesis-II). iCORPP with complete and accurate knowledge, corresponding to the curve with circle markers, produces the best performance, while the agent is able to recover from "inaccurate" knowledge by asking more clarification questions (corresponding to the higher identification cost).

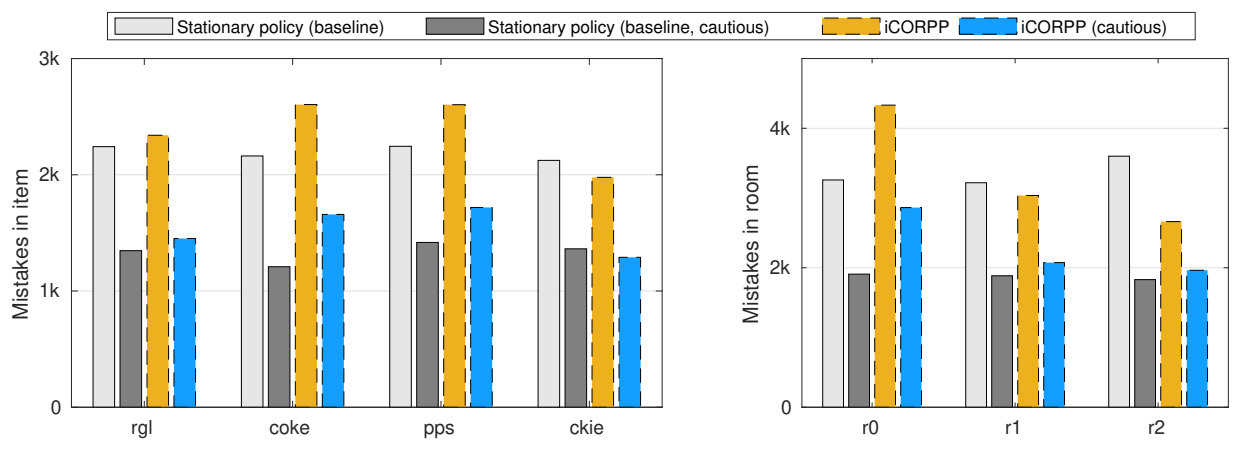

**Figure 12:** iCORPP enables the robot to fine-tune its behavior in delivering different items to different rooms. The x-axis and y-axis correspond to the *incorrect* deliveries and the number of mistakes (over $100k$ trials). It should be noted that reducing the number of mistakes is not the goal in this experiment, as it can be easily achieved by increasing the penalty of failures in dialog (Zhang and Stone, 2015).

For completeness, we evaluated the performance of pure logical-probabilistic reasoning (about $\mathcal{K}^S$), where information gathering actions are not allowed. The robot uses all knowledge to determine the most likely delivery request (in case of a tie, it randomly chooses one from the most likely ones). The reasoning takes essentially no time and the average accuracy is only 0.193, which is significantly lower than strategies that involve POMDP-based active information gathering.

[6] A part of Figure 11 appeared in a previous article that focused on a robot software architecture, called Building Wide Intelligence (BWI), for human-robot interaction in general (Khandelwal et al., 2017).

In the next experiment, we aim at evaluating Hypothesis-III, i.e., iCORPP enables to fine-tune agent behaviors at a level, where a comparable hand-coded controller requires a prohibitively large number of parameters. The spoken dialog system includes four items, three rooms and two persons, resulting in a relatively small state space. We give the robot the ontology of items, as shown in Figure 9. The hidden shopping request was randomly selected in each trial. Speech recognition errors are modeled, e.g., 0.8 accuracy in recognizing answers of confirming questions and a lower accuracy for general questions (depending on the number of that sort's objects).

Figure 12 shows the numbers of mistakes made by the robot. In the default and cautious versions of iCORPP, the values of $[R^+, R^-]$ are $[20, -20]$ and $[30, -30]$ respectively. The first observation is that the baseline method that builds on a stationary POMDP-based policy makes no difference in either item (**Left**) or room (**Right**), because it does not reason about the reward system—incorrect deliveries are not differentiated and all receive the same penalty. In contrast, both versions of iCORPP enable the robot to behave in such a way that the robot makes the fewest mistakes in `cookie` (**Left**) and room `r2` (**Right**). Such behaviors match our expectations: `cookie` is "very different" from the other three items and `r2` has the greatest distance from the shop, so the robot should make effort to avoid delivering `cookie` (or delivering to `r2`) when that is not requested. Without iCORPP, to achieve such fine-tuned behaviors, there will be 600 parameters in the reward function need to be hand-coded, which is impossible from a practical point of view, which supports Hypothesis III.

The second (side) observation from Figure 12 is that one can adjust the robot's "cautious level" (overall success rate) by tuning the values of $[R^+, R^-]$. Comparing the default versions of stationary policy and iCORPP, we see the default iCORPP producing more mistakes, because our implementation of iCORPP gave partial credits to less severe mistakes. To avoid introducing extra mistakes, one can adjust $[R^+, R^-]$, as was done in our previous work (Zhang and Stone, 2015), or instead give extra bonus rewards to mistake-free trials.

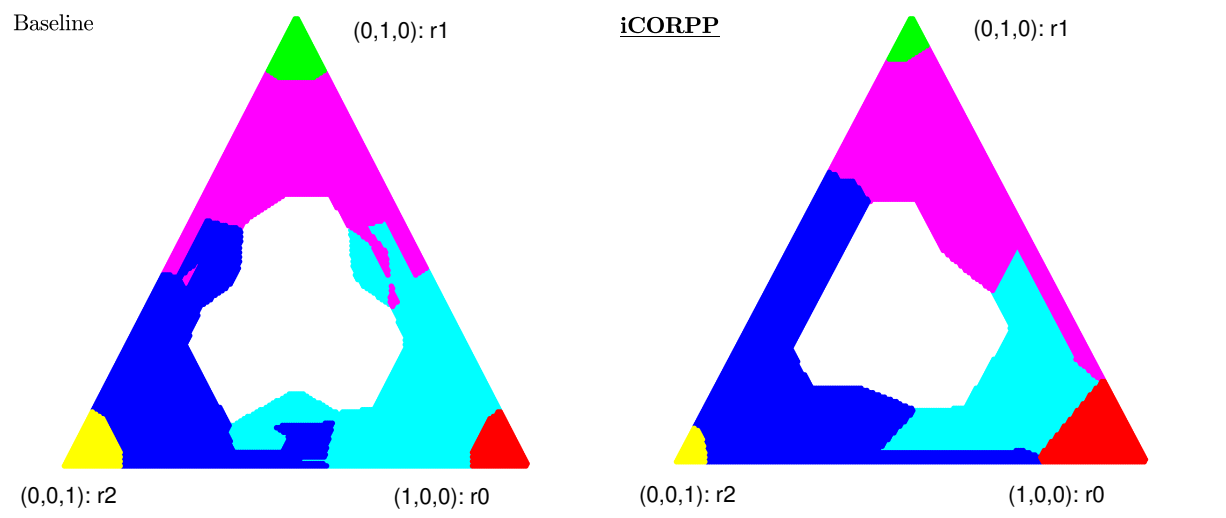

**Figure 13:** A visualization of a POMDP-based policy where all wrong deliveries are equally penalized (baseline), and the iCORPP policy where the reward function is computed via logical-probabilistic reasoning. In this experiment, the person wants to deliver to one of the three rooms. Each point in the two subfigures corresponds to a belief. Each color corresponds to an action: white corresponds to the general question of "which room to deliver"; the colors in the corners correspond to delivery actions; and the remaining three colors correspond to confirming questions.

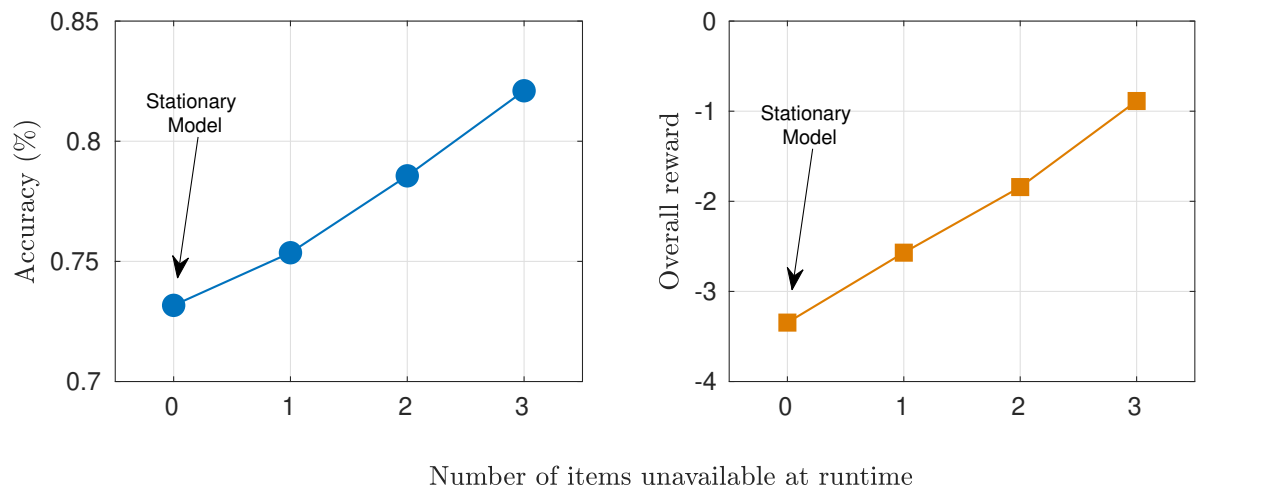

**Figure 14:** iCORPP performs increasingly well in accuracy and overall reward in the shopping task when more items are known to be unavailable: the baseline (stationary model) corresponds to the left ends of the two curves, where the baseline model has to include all items.

To better understand the robot's behavior (specifically, the **Right** of Figure 12), we manually remove the uncertainties in `item` and `person` in the initial belief, and visualize which action the POMDP policy suggests given different initial beliefs in `room`. In the **Right** of Figure 13, we see the robot is relatively more cautious in delivering to `r1` and `r2` (the green and yellow areas in the top and left corners are smaller than the red one in the right), because rooms `r1` and `r2` are relatively far away from the shop, as shown in Figure 5(a). It is very difficult to achieve such fine-tuned behaviors from hand-coded models, because of the prohibitively large number of parameters in the reward system. In contrast, iCORPP reasons with logical-probabilistic knowledge to construct the transition and reward systems (Section 5.3).

Figure 14 shows the results of the shopping task when exogenous changes are added: items can be temporarily unavailable. iCORPP dynamically constructs POMDPs: when items are known to be unavailable, states of these items being requested and actions of delivering these items are removed from the POMDP. For instance, when three items are unavailable, the numbers of states and actions are reduced from $(37, 50)$ to $(18, 29)$. As a result, iCORPP performs better in both accuracy and overall reward (y-axes in Figure 14) when more items are known to be unavailable (x-axes in Figure 14). In contrast, the baseline, using a static POMDP, must include all items (assuming no item unavailable), because it cannot adapt to exogenous changes. So the baseline's performance corresponds to the left ends of the two curves. Results shown in Figure 14 support that iCORPP enables the robot to adapt to exogenous domain changes, whereas stationary policies do not (Hypothesis-IV). A demo video is available at this link: `http://youtu.be/2UJG4-ejVww`

## 6. Conclusions and Future Work

This article introduces a novel algorithm called iCORPP that uses commonsense reasoning to dynamically construct (PO)MDPs for scalable, adaptive robot planning. iCORPP uses declarative language P-log for logical-probabilistic knowledge representation and reasoning, and uses probabilistic graphical models, such as (PO)MDPs, for probabilistic planning. This article, for the first time, enables robot behaviors to adapt to exogenous domain changes without including these exogenous attributes in probabilistic planning models. iCORPP has been evaluated both in simulation and on a real robot. We observed significant improvements comparing to competitive baselines (including hand-coded action policies), based on experiments using problems of mobile robot navigation and spoken dialog systems in an office environment.

***Applicability of iCORPP:***

iCORPP decomposes a problem of sequential decision-making under uncertainty into two subproblems of commonsense reasoning and probabilistic planning that respectively focus on the "curse of dimensionality" and the "curse of history" – as elaborated in (Kurniawati et al., 2011). In this process, commonsense reasoning aims to understand the current state and dynamics of the world, and probabilistic planning focuses on task-oriented action selection toward goal achievement. Therefore, iCORPP significantly reduces the complexity of (PO)MDP planning compared to its one-shot solution, while enabling robot behaviors to adapt to exogenous changes.

Consider a mobile robot navigation domain from Section 5.2 that includes only thirty positions, five weather conditions, and three times. There are human walkers who can probabilistically disrupt the robot's navigation actions. One can naively enumerate all combinations of attribute values (Boutilier et al., 1999). The enumeration produces a large number of states,

$$N = |Loc| \cdot 2^{2 \times |Loc|} \cdot |Weather| \cdot |Time| \cdot |Term|$$

where $Loc$, $Weather$, and $Time$ are sets of locations, weathers, and times respectively. The value of $|Term|$ is 2, where $term \in Term$ can be true or false, used for identifying the end of an episode. Back to this small domain, naive enumeration produces more than 2ˆ69 states, making it impossible to produce a meaningful policy in a reasonable amount of time. In comparison, the MDP constructed by iCORPP includes only 60 states ($|Loc| \cdot |Term|$), and can be readily handled using off-the-shelf planning systems.

*Default Reasoning:* We use defaults when a complete world model is unavailable or reasoning with such models requires prohibitive computing resources. Continuing the above-mentioned navigation example, it is possible that the *Weather* variable's value could not be observed in the environment for reasons such as sensor failures. In that case, the robot has at least the two options: 1) reasoning with defaults (e.g., assuming the weather is sunny), and 2) inferring the weather based on fully observable evidence. For instance, people holding an umbrella and wet ground can be evidence of rainy days. However, the introduction of new domain variables and their interdependencies increases the complexity of at least the reasoning subproblem. For practical reasons, iCORPP practitioners might want to assign default values to avoid the "curse of dimensionality" in reasoning, where the defaults can be "defeated" when their corresponding values can be extracted from the real world. Generally, there is the trade-off between model completeness and computational tractability, and default reasoning (well supported by P-log) provides a realization of such trade-offs.

*Inapplicability:* iCORPP is inapplicable when reasoning or planning is unnecessary. In the extreme, when there are no exogenous variables, reasoning becomes unnecessary; when there are no endogenous variables, planning becomes unnecessary. There is also a "gray area", where iCORPP can be less effective. For instance, when the provided knowledge is generally useful but less relevant to the current task, the reasoning results from iCORPP will not be useful for action selections in the planning steps. The evaluation of iCORPP's effectiveness in general is difficult, but case-by-case analyses can be conducted by iCORPP practitioners.

*Closed-World Assumption (CWA):* There is usually a CWA in logical reasoning, which we adopt in Algorithm 1 (Line 4), meaning that what is not currently known to be true is believed false. CWA ensures that every variable (exogenous or endogenous) has a value in each possible world. More precisely, under CWA, each entry in $w^{cplt} := [v_0^{en}, v_1^{en}, \cdots]$ in Line 4 of Algorithm 1 has a value. Without CWA, it becomes an open question how to deal with statements on variables that are neither true nor false, making it very hard to specify the state space. The default reasoning capability of ASP facilitates our implementation of CWA, and the default values can be easily defeated by facts when available. The general applicability of iCORPP under Open-World Assumption (OWA) is beyond the scope of this article.

***Future Work:***

There are a number of ways to make further progress in this line of research. First, learning is not incorporated into iCORPP. We are currently investigating improving iCORPP by using supervised learning to help estimate the current world state (Amiri, Shirazi and Zhang, 2020) and using model-based reinforcement learning to update declaratively-represented world dynamics (Lu et al., 2020). The logical-probabilistic knowledge base is manually encoded, whereas data mining algorithms (Han, Kamber and Pei, 2012) and publicly available knowledge bases, such as Open Mind Common Sense (OMCS) (Singh, Lin, Mueller, Lim, Perkins and Zhu, 2002) and ConceptNet (Speer, Chin and Havasi, 2017), can be used to augment the knowledge base. Second, other reasoning and planning paradigms can be used to further improve the system performance. For instance, Markov Logic Networks (Richardson and Domingos, 2006) and Probabilistic Soft Logic (Kimmig et al., 2012) have well maintained systems that can potentially improve the reasoning component of iCORPP. Third, iCORPP assumes the current world state is either fully observable or all variables are partially observable. There is the potential of applying iCORPP to domains with more complex observabilities, e.g., mixed observability as investigated in our recent research (Amiri, Wei, Zhang, Sinapov, Thomason and Stone, 2018). Given that robots' long-term autonomy capabilities continue to improve, we can conduct more experiments to evaluate the performance of iCORPP under different conditions. For instance, robots with relatively weak perception capabilities can better benefit from iCORPP's reasoning capability, whereas iCORPP's planning capability (for active perception) is relatively more important in highly dynamic environments. Such hypotheses can be evaluated using real robots in the future.

The recent advances in pretrained large language models (LLMs), e.g., GPT-3 (Ouyang, Wu, Jiang, Almeida, Wainwright, Mishkin, Zhang, Agarwal, Slama, Ray et al., 2022), ChatGPT (OpenAI, 2023), and LLaMA (Touvron, Lavril, Izacard, Martinet, Lachaux, Lacroix, Rozière, Goyal, Hambro, Azhar et al., 2023), have reshaped the landscape of AI. iCORPP assumes that the domain knowledge is provided by a human, and the goals are provided in a rule-based formal way. LLMs have made it possible to remove those assumptions. For instance, our recent work has demonstrated that LLMs allow the planning goals to be specified in natural language (Liu, Jiang, Zhang, Liu, Zhang, Biswas and Stone, 2023), and knowledge can be extracted from LLMs to assist classical planning (Ding, Zhang, Amiri, Cao, Yang, Kaminski, Esselink and Zhang, 2023a; Ding, Zhang, Paxton and Zhang, 2023b). Whether stand-alone LLMs can be used for planning is still an open question. While there are successes on LLM-based planning demonstrated in recent literature, e.g., (Huang, Abbeel, Pathak and Mordatch, 2022; Driess, Xia, Sajjadi, Lynch, Chowdhery, Ichter, Wahid, Tompson, Vuong, Yu et al., 2023), there are many domains where LLM-based planning systems do not perform well (Valmeekam, Marquez, Sreedharan and Kambhampati, 2023). Instead of directly addressing the long-horizon challenge, iCORPP provides one way of decomposing sequential decision-making tasks into the two subtasks of reasoning and planning. In this article, the decomposition strategy is manually defined, but LLMs can potentially introduce new approaches for generating more manageable subtasks (Wei, Wang, Schuurmans, Bosma, Xia, Chi, Le, Zhou et al., 2022), which may lead to very interesting future research.

## Acknowledgements

This work has taken place in the Autonomous Intelligent Robotics (AIR) group at SUNY Binghamton and in the Learning Agents Research Group (LARG) at the Artificial Intelligence Laboratory, The University of Texas at Austin. AIR research is supported in part by grants from the NSF (IIS-1925044), Ford Motor Company, OPPO, and SUNY Research Foundation. LARG research is supported in part by the National Science Foundation (FAIN-2019844, NRT-2125858), the Office of Naval Research (N00014-18-2243), Army Research Office (E2061621), Bosch, Lockheed Martin, and Good Systems, a research grand challenge at the University of Texas at Austin. The views and conclusions contained in this document are those of the authors alone. Peter Stone serves as the Executive Director of Sony AI America and receives financial compensation for this work. The terms of this arrangement have been reviewed and approved by the University of Texas at Austin in accordance with its policy on objectivity in research.

## CRediT authorship contribution statement

**Shiqi Zhang:** Conceptualization of this study, Methodology, Software. **Piyush Khandelwal:** Conceptualization of this study, Methodology, Software. **Peter Stone:** Conceptualization of this study, Methodology.